\documentclass{article} 
\usepackage{iclr2026_conference,times}

\usepackage{amsmath,amsfonts,bm}

\def\eqref#1{equation~\ref{#1}}

\def\1{\bm{1}}

\DeclareMathAlphabet{\mathsfit}{\encodingdefault}{\sfdefault}{m}{sl}
\SetMathAlphabet{\mathsfit}{bold}{\encodingdefault}{\sfdefault}{bx}{n}

\usepackage{hyperref}
\usepackage{url}
\usepackage{graphicx}
\usepackage{booktabs}
\usepackage{multirow}
\usepackage{array}
\usepackage{makecell}

 \title{Investigating Data Interventions for \\ Subgroup Fairness: An ICU Case Study}

\author{Erin Tan \\
University of California, Berkeley \\
\And
Judy Hanwen Shen \\
Stanford University \\
\AND
Irene Y. Chen \\
University of California, Berkeley \\
}

\iclrfinalcopy 
\begin{document}

\maketitle

\begin{abstract}
In high-stakes settings where machine learning models are used to automate decision-making about individuals, the presence of algorithmic bias can exacerbate systemic harm to certain subgroups of people. These biases often stem from the underlying training data. In practice, interventions to ``fix the data" depend on the actual additional data sources available---where many are less than ideal. In these cases, the effects of data scaling on subgroup performance become volatile, as the improvements from increased sample size are counteracted by the introduction of distribution shifts in the training set. In this paper, we investigate the limitations of combining data sources to improve subgroup performance within the context of healthcare. Clinical models are commonly trained on datasets comprised of patient electronic health record (EHR) data from different hospitals or admission departments. Across two such datasets, the eICU Collaborative Research Database and the MIMIC-IV dataset, we find that data addition can both help and hurt model fairness and performance, and many intuitive strategies for data selection are unreliable. We compare model-based post-hoc calibration and data-centric addition strategies to find that the combination of both is important to improve subgroup performance. Our work questions the traditional dogma of “better data” for overcoming fairness challenges by comparing and combining data- and model-based approaches. 
\end{abstract}

\section{Introduction}
When machine learning algorithms are used to guide real-world decision making, they are at risk of reproducing disparities found in the training data~\citep{Chen2018WhyIM, doi:10.1089/big.2016.0047, angwin2022machine}. The majority of technical advancement has focused on model interventions that remedy the learned disparities of the model~\citep{hebert2018multicalibration, agarwal2018reductions}. Many have also called for sociotechnical approaches such as abstaining from using AI~\citep{Solarova2022ReconsideringTR, Dressel2018TheAF, binns2018s} or system approaches such as collecting more data~\citep{Chen2018WhyIM,  Drukker2023TowardFI, chen2023ethics}.   

``Improving the data," in particular, has been frequently suggested as a vague, catch-all solution to the problem~\citep{arora2022, HUANG2024104622}.  However, there is a gap between broad suggestions of data remedies and what is effective in practice. There are three main challenges that practitioners face when translating between the promise of better data and practical methods of data addition. First, the best method of addition is task-dependent; the success of an approach in one method does not guarantee success on another dataset~\citep{Rommel_2022}. Second, we are restricted by the quality of the available data. The limitations of the dataset do not exist because practitioners \textit{choose} to use suboptimal data, but because better data is simply unavailable. Third, the effect of data changes on subgroups as well as the general population must be considered. 

Due to the first challenge---that data limitations are task dependent---we narrow our focus on the case study of intensive care unit (ICU) data in particular. Bias in healthcare is particularly consequential, as disparities may translate into unequal treatment quality~\citep{mittermaier2023bias}, delayed or missed diagnosis~\citep{CheXclusion_2020, Underdiagnosis_2021}, or increased costs of care~\citep{doi:10.1126/science.aax2342}. Moreover, we find clinical EHR datasets to be well-representative of the second and third challenges described above. To combat the lack of quality data, EHR datasets are often aggregated across multiple sources (e.g. multiple hospitals or admission departments); this strategy allows for larger dataset sizes, yet effective methods for achieving better data quality remain unclear. Finally, EHR datasets contain heterogeneous patient subgroups, so the notion of better data may depend on the subgroup in question. Grounding our work in this domain allows us to better isolate the impact of specific data interventions, although we emphasize that our methods can be applied to any multi-source dataset.


Our work focuses on the specific goal of improving subgroup performance through data interventions. Specifically, our contributions are:  
 
\begin{itemize}
    \item We review the landscape of data interventions for fairness, identifying a gap between recommended data solutions and available algorithms.
    \item We evaluate and compare existing data interventions for subgroup performance and find that they are not effective across two datasets---the eICU Collaborative Research Database and MIMIC-IV. We identify mean discrepancy as a key barrier to success for data addition.
    \item We compare data-centric approaches with post-hoc model calibration and identify a hybrid approach best improves subgroup performance. 
\end{itemize}

\section{Related Work}





\paragraph{Data Perspectives on Fairness}
Pre-processing fairness interventions address biases that exist within the training data~\citep{PESSACH2021115667, zhioua2025originssamplingbiasimplications}. These biases may arise as a result of systematic faults in data collection, where collection methods inherently under-represent certain groups or promote differences in data quality, labeling procedure, or missingness. When biases exist in the data, they are near-impossible to correct using in-process or post-processing interventions alone~\citep{Chen2018WhyIM, rolf2021representation}. 


``Adding more data" does not account for practical realities---in most cases, we can assume that in-distribution data has already been exhausted. Additionally, new data may be drawn from a biased distribution, such that adding more samples from the same distribution can amplify those disparities~\citep{zhioua2025originssamplingbiasimplications}. 
Adding data from external data sources yields unpredictable outcomes, as the counteracting effects from increased sample size and introduction of distribution shift work against one another~\citep{pmlr-v252-shen24a, wang2022robustfairnessevaluatingsustaining}. Fairness under distribution shifts is generally evaluated by quantifying the worst-case performance within a set of potential shifts~\citep{JMLR:v25:23-0739, wang2022robustfairnessevaluatingsustaining}. Distributionally-robust optimization (DRO) uses this notion by directly optimizing worst-case subgroup performance under distribution shift during training~\citep{rahimian2022frameworks}. Parallel work in invariant learning aims to discover more generalizable patterns in data which remain stable across environments, but these are in-process methods~\citep{arjovsky2020invariantriskminimization, 10.1007/978-3-030-58526-6_44}.

\paragraph{Data Interventions}
Data interventions for subgroup fairness are most commonly evaluated in the single-source setting. In general, the goal of all these data interventions is to mitigate the problem of imbalanced class distributions, a common cause of sample size bias and under-representation bias~\citep{zhioua2025originssamplingbiasimplications}. These include sampling-based techniques, such as over-sampling from minority classes~\citep{pmlr-v177-idrissi22a, Chawla_2002} or under-sampling from majority classes~\citep{LIN201717}. Standard over-sampling selects samples from the minority group to duplicate, which helps in adding more data, but does not offer any new information and is prone to overfitting the minority distribution. Newer methods use generative models, such as variational autoencoders, to add synthetic data, which work better on high-dimensional data than sampling~\citep{wan2017vae}.

Another approach to balancing subgroups is reweighing, which assigns importance weights to training instances, such that minority examples are given greater influence over the loss function to equalize the contributions of all subgroups ~\citep{5360534, article}.


\paragraph{Calibration}

Calibration is one way to combat algorithmic bias---for all individuals receiving a predicted probability of $p$, ensure that a $p$ fraction of them are positive samples in actuality. When we satisfy this constraint for all subgroups across some sensitive attribute (e.g. race or gender), we ensure that the model predictions give meaningful estimates of uncertainty for all individuals.

Generally, model calibration is performed in post-processing by learning a $1$-dimensional transformation function $\phi: [0, 1] \rightarrow [0,1]$ mapping the uncalibrated model predictions $f(x)$ to their calibrated form $\phi(f(x))$. For example, Isotonic Regression~\citep{10.1145/775047.775151} is a non-parametric calibration technique which learns a monotonically non-decreasing $\phi$. 
We use Isotonic Regression in our experiments to compare pre-processing and post-processing interventions.

\section{Problem Setup}
Our goal is to improve the performance of a target subgroup using data interventions. In a real-world setting, our primary options for increasing data are to either 1) generate synthetic samples or 2) add out-of-distribution samples from external sources. In this work, we focus on the latter. In this section, we establish a formal framework for examining the impact of multi-source data addition on model fairness across subgroups. 

\begin{figure}[t]
\begin{center}
\includegraphics[width=0.9\linewidth]{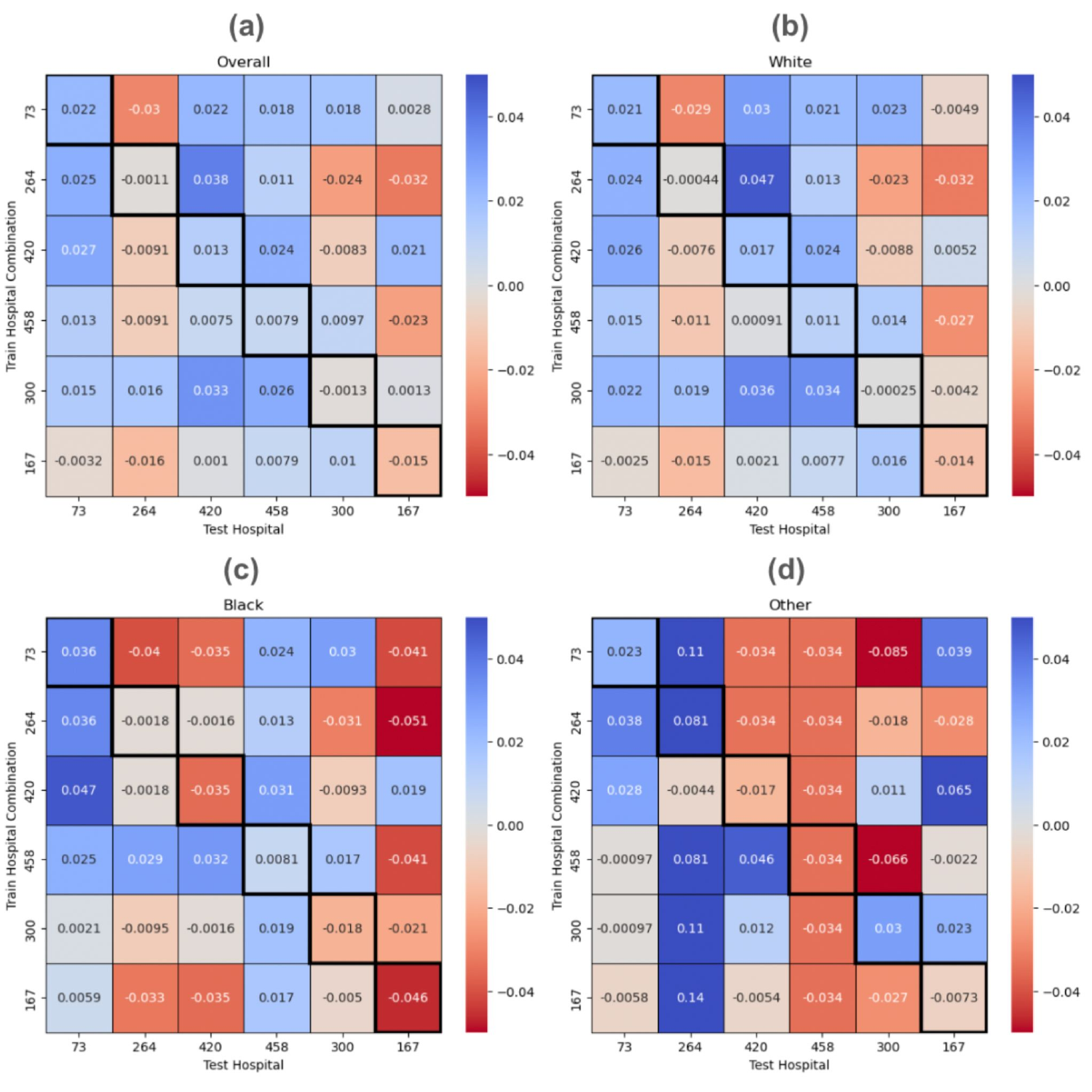}
\end{center}
\caption{\small Change in Overall and subgroup-level accuracy after \textsc{Whole-Source} data addition (Logistic Regression). The change in Overall performance (a) does not reflect equally upon changes in performance across subgroups. For example, we observe that adding data from any source to Target Hospital 458 improves overall accuracy. While this change is reflected across the White (b) and Black (c) subgroups, the Other (d) subgroup experiences near-uniform decreases in test performance as a result of data addition. Scaling data directly from the Target Hospital (diagonal of each plot) generally yields improvements in overall and subgroup accuracy, but is typically not the \textit{best}-performing data addition choice.}
\label{fig:data_addition_heatmaps}
\end{figure}

We begin with a model $f_{D_{\text{train}}}(D_{\text{test}}): X \mapsto \{0, 1\}$ that is trained on dataset $D_{\text{train}} = \{(x_i, y_i)\}_{i=1}^{n_{\text{train}}}$ and evaluated on $D_{\text{test}} = \{(x_i, y_i)\}_{i=1}^{n_{\text{test}}}$, where we consider a binary target variable: $y \in \{0, 1\}$. 
We also have some discrete sensitive attribute $A$ within our feature set $X$ (e.g. race) which we can use to stratify our datasets into subgroups (e.g. by race: Asian, Black, White). We identify some target subgroup $a' \in A$ whose performance we would specifically like to improve, and use $D_{\text{test}}^{a'}$ to denote the subset of $D_{\text{test}}$ where $A = a'$. Let $0 < t < 1$ represent the prediction threshold such that if the probability outputted by $f_{D_\text{train}}(X_i) \ge t$, then the predicted label $\hat{y} = 1$, and $0$ otherwise. Then, for any model $f$ and subgroup $a$ we define the empirical subgroup accuracy as 

\begin{equation}
    \text{Acc}(f, a) = \frac{1}{|D_{\text{test}}^a|} \sum_{(x, y) \in  D_{\text{test}}^a} 1 - |y - \mathbf{1}[f(x) \geq t]| \label{eq:acc}
\end{equation}

In our setup, we assume that the model class and complexity are fixed, such that our only available levers for improving subgroup accuracy are to modify $D_{\text{train}}$ by adding data from external sources, denoted $\mathcal{S}_1, \mathcal{S}_2, \dots, \mathcal{S}_n$. Each source $i$ has some available dataset, denoted $D_{i}$, which is drawn from the source distribution $\mathcal{S}_i$.  Formally, our goal is to find a dataset $D_* = D_{\text{train}} \cup \bigcup_{i=1}^n z_iD_i, \quad \text{where } \textbf{z} \in \{0, 1\}^n$ which optimizes the subgroup accuracy over fixed target subgroup $a'$. $D_*$ is assembled by adding datasets from zero or more available sources to $D_{\text{train}}$. One way to compose $D_*$ is by greedily selecting $i^*$, the next source to add from.
\begin{equation}
    i^* = \arg \max_{i} \ \ \ \text{Acc}(f_{D_{\text{train}} + D_i}(D_{\text{test}}^{a'}))
\end{equation}

Our goal with this work is to better understand the characteristics of datasets that are most informative for composing $D_*$ efficiently.

\section{Experiments} \label{sec:experiments}

\subsection{Datasets} 
We evaluate our methods on two datasets: the eICU Collaborative Research Database~\citep{pollard2018eicu, PhysioNet} and the MIMIC-IV dataset~\citep{Johnson2023, PhysioNet} and use patient ethnicity as the sensitive attribute of interest. 

The eICU dataset contains 200,859 patient unit encounters in 208 hospitals across the US. We include 12 hospitals, each with over 2000 patients. Each of these hospitals is treated as its own data source.\footnote{The demographic information for eICU can be found in Table \ref{tab:eicu-data} in the Appendix.} MIMIC-IV contains 546,028 patient admissions from over 265,000 unique patients to various departments in the facility, including the emergency unit, intensive care unit, and urgent care. 
We treat each admission department as a different data source.\footnote{The demographic information for MIMIC can be found in Table \ref{tab:mimic-data} in the Appendix. We also include additional results from experiments using the MIMIC-IV dataset for the task of in-hospital mortality prediction in the Appendix.} 

\subsection{Models and Evaluation} 
We investigated three models prominent in prior work \cite{vandewaterYetAnotherICUBenchmark2024}: a Logistic Regression (LR) model, a tree-based Light Gradient Boosting Machine (LGBM) model, and a Long Short-Term Memory (LSTM) model. 
We used the same set of features for all models and used 5-fold validation for all results reported. 

We evaluate subgroup accuracy and AUC (reported in the Appendix). Accuracy is defined in Equation \ref{eq:acc}. We additionally define subgroup AUC over $D_{\text{test}}$, which can be partitioned into $D_0$ and $D_1$, representing the subsets of negative and positive samples, respectively:

\begin{equation}
    \text{AUC}(f, a) = \frac{1}{|D_0^{a}| |D_1^{a}|} \sum_{x_i \in D_0^{a}, x_j \in D_1^{a}} \mathbf{1}[f(x_j) > f(x_i)]
    \label{eq:auc}
\end{equation}

\subsection{Data Addition Experiments}
To test our data addition heuristics, we performed three types of experiments, detailed in this section.
\begin{itemize}
    \item \textsc{Baseline}: For each source and model class, we train on $1000$ samples and test on $400$ samples. Change in performance is measured with respect to the \textsc{Baseline}. 
    \item \textsc{Whole-Source}: For each data source, we train a model on the combined training set of the $1000$ samples from the test data source, as well as $1000$ samples from the additional data source. As a control, we also train a model on $2000$ samples for each test source.
    \item \textsc{Subgroup-Level}: To the existing test data source of $1000$ data points, we add only a subset of a source $D_i^a$ without adding all other subgroups in $D_i$. Additional samples are capped at $1000$, although the vast majority of subgroups are smaller.
\end{itemize}

\section{Limitations of Existing Data Addition Methods} \label{sec:naive-data-addition}

We motivate the need for informed data selection heuristics by exploring the complex realities of fairness under data addition. Figure \ref{fig:data_addition_heatmaps} shows the change in overall and subgroup accuracies after \textsc{Whole-Source} data addition compared to the \textsc{Baseline}.

Our key observation is that the effects of data addition vary across subgroups. For example, for Target Hospital $458$, addition of any data source improves the Overall, White, and Black accuracies but worsens outcomes for the Other subgroup. Similarly, for Target Hospital $420$, predictor accuracy for Overall and White subgroups improved by data addition with the Black and Other subgroups sometimes suffered significantly worsened accuracies. This is even the case where more data from the \textit{same} source is added.

\begin{figure}
    \centering
    \includegraphics[width=0.6\linewidth]{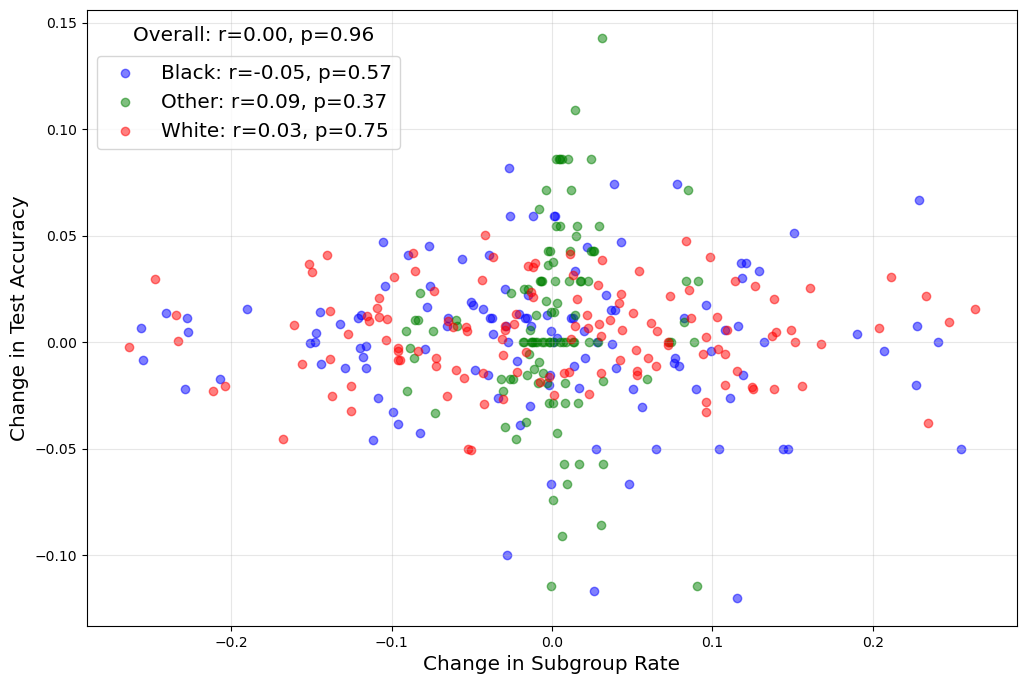}
    \caption{\small Change in subgroup ratio vs. Change in subgroup test accuracy after \textsc{Whole-Source} data addition on the eICU Dataset.}
    \label{fig:subgroup-rate}
\end{figure}
These findings show that it is 1) necessary to view the effects of data interventions (e.g. addition) at a subgroup level and 2) that it is not always beneficial to add all available data into your training set when subgroup performance matters. We need to make careful data decisions by considering the effects on different subgroups, but it remains unclear how to improve subgroup performance in this setting. In the following subsections, we investigate several common data selection strategies, showing that they are not effective in improving subgroup performance in our settings.

\paragraph{Data Balancing}
The first set of informed data selection strategies we evaluated seeks to improve the presence of the target subgroup within the training dataset~\citep{cohen2021}. Traditionally, subgroup balancing is performed either by oversampling from minority classes~\citep{Chawla_2002} or undersampling majority classes~\citep{LIN201717}. 

The intuition for data balancing via selection is that, for some minority group $a'$, we select the dataset $D_{i_*}$ that maximizes the representation of $a'$. Let $|D_i|$ denote the total number of samples available from source $\mathcal S_i$. We investigate two different selection criteria to perform data balancing via addition:
\begin{enumerate}
    \item Adding sources with the highest proportion of samples from the target subgroup
    $$i_* = \arg \max_i \frac{|D_i^{a'}|}{|D_i|}$$
    \item Adding sources with the greatest number of samples from the target subgroup
    $$i_* = \arg \max_i |D_i^{a'}|$$
\end{enumerate}

In Figure \ref{fig:subgroup-rate}, we plot the changes in subgroup ratio after \textsc{Whole-Source} data addition against the change in subgroup performance. The ratio of a subgroup within a dataset is computed as $\text{Ratio}(D, a) = \frac{|D^a|}{|D|}$, so that we define the change in subgroup ratio as $\Delta \text{Ratio}(D_{\text{train}}, D_{\text{added}}, a) = \text{Ratio}(D_{\text{train} \cup \text{added}}, a) - \text{Ratio}(D_{\text{train}}, a)$. 

We did not observe a statistically significant relationship in  \ref{fig:subgroup-rate} across any subgroup; increasing the ratio of a subgroup does not reliably improve subgroup performance in isolation.

\begin{figure*}[t]
    \centering
    \includegraphics[width=1\linewidth]{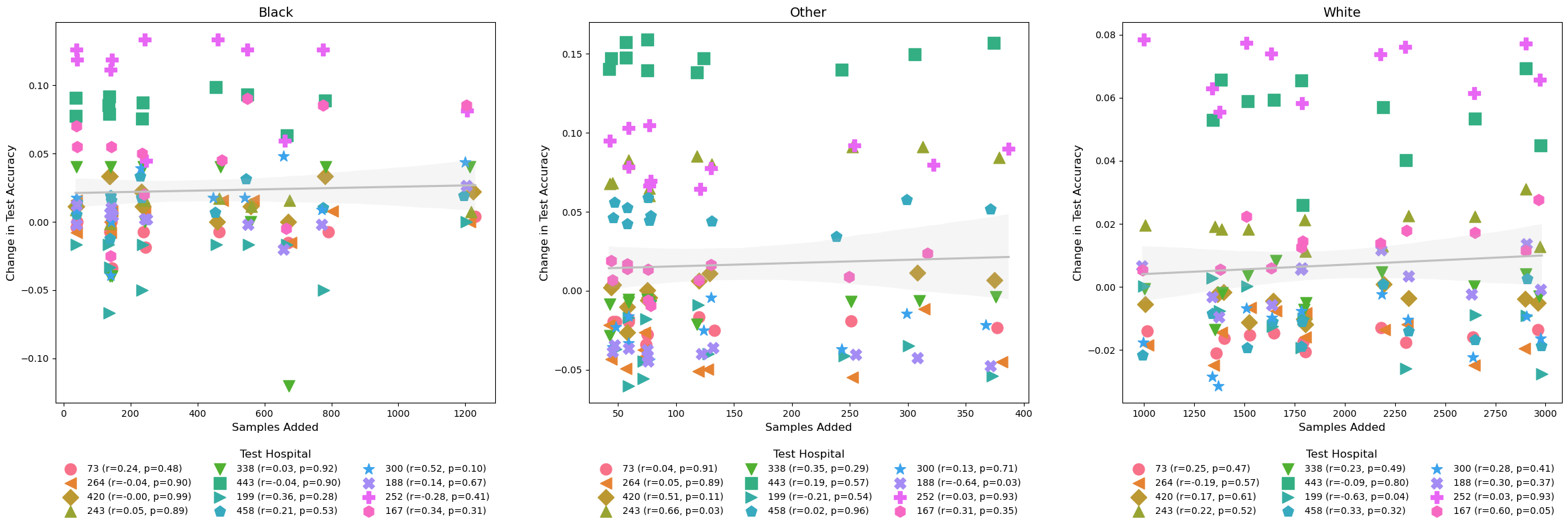}
    \caption{\small Change in subgroup accuracy as a function of samples added in \textsc{Subgroup-Level} data addition (see Section \ref{sec:experiments} for details). Across nearly all combinations of subgroups and Test Hospitals, we find that adding more samples does not necessarily lead to larger performance gains. These visualizations lead to the conclusion that naive subgroup balancing is an uninformative data selection heuristic.}
    \label{fig:samples-added}
\end{figure*}

To test the second heuristic---choosing the source with the most available samples for the target subgroup---we examine the change in performance after \textsc{Subgroup-Level} data addition. The results are presented in Figure \ref{fig:samples-added}, where we find that the quantity of available subgroup data within a source is not meaningful for determining how data addition will steer subgroup performance. 

Through our first two data selection methods---selecting the sources with the greatest number and highest ratio of samples from the test subgroup, respectively---we found that simply increasing the representation of samples from the test subgroup does not necessarily yield positive results if the data itself is uninformative about the target.

\paragraph{Distribution Matching}
One critical observation from Figure \ref{fig:samples-added} is that it suggests it is more important to understand the \textit{test} distribution rather than the just the added source distribution, as some test sources generally benefit from any type of data scaling (e.g. Hospital 443, 252), whereas others exhibit minimal or negative effects from any out-of-distribution scaling (e.g. Hospital 199, 73). Thus, the next data selection method we evaluate chooses the sources which are most similar to the test distribution. Intuitively, this should yield desirable outcomes because we minimize the amount of distribution shift while still scaling the number of training samples.

We employ the subgroup similarity score proposed in~\citeauthor{pmlr-v252-shen24a} as a heuristic to quantify the distance between two source distributions. Given that $D_{\text{test}}^a \sim \mathcal S_{\text{target}}^a = P$ and $D_i \sim \mathcal{S}_i = Q$ for any external source $i$, we derive a binary predictor $s_{PQ}: X \rightarrow [0, 1]$ where datapoints belonging to distribution $P$ are assigned label $1$ and points in $Q$ are assigned label $0$. Intuitively, if $P$ and $Q$ are similarly distributed, $s$ should output probabilities around $0.5$ (random guessing) on an empirical test set containing points from $P$. If $P$ and $Q$ are very different, $s_{PQ}$ should output probabilities closer to $1$. Thus, we can use as a heuristic for distributional similarity the expected value of the predictor probability on $P$ while assuming a uniform probability for the empirical samples from $P$:
\begin{equation}
    \textsc{Score XY} = \mathbb{E}_{(x, y) \in P}[s_{PQ}(x, y)]\label{eq:score-xy}
\end{equation}

Applying our similarity score function to the subgroup case, we define the subgroup similarity score for a shared subgroup $a$ between two sources to be:
\begin{equation}
    \textsc{Score XY}_a = \mathbb{E}_{(x, y) \in P}[s_{PQ}(x, y) | (x, y) \in a]\label{eq:subgroup-score-xy}
\end{equation}

\begin{figure*}
    \centering
    \includegraphics[width=1\linewidth]{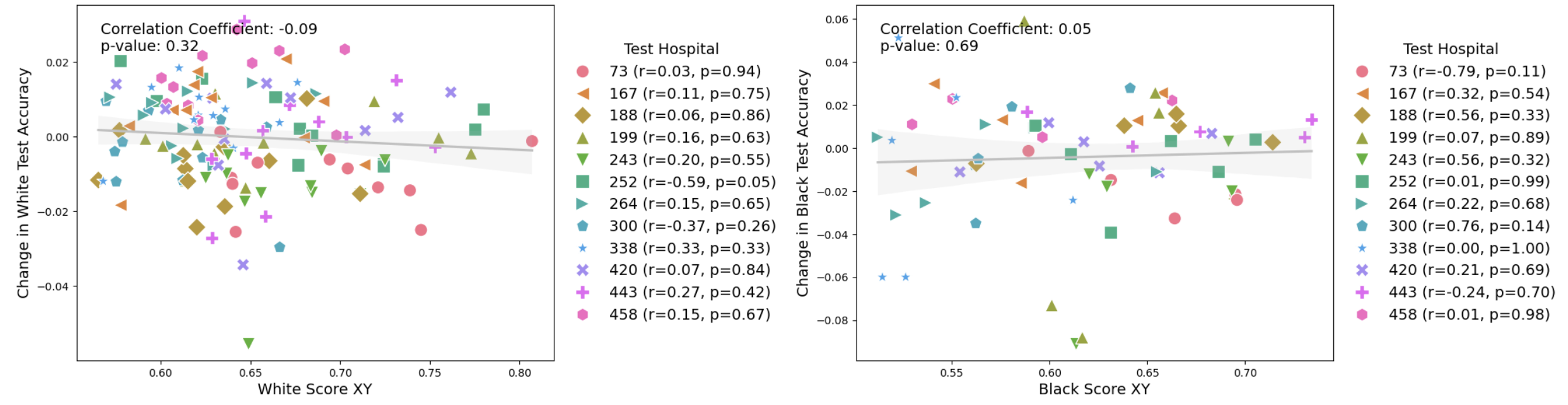}
    \caption{\small Subgroup similarity score (using features and labels) vs. Change in subgroup test accuracy for the White (left) and Black (right) subgroups (eICU). The scores are computed using only the patients from the target subgroup in each source. These performance changes result from \textsc{Subgroup-Level} data addition. We do not observe statistically significant correlations in any Test Hospital for either subgroup.}
    \label{fig:subgroup-score}
\end{figure*}

We again consider the results from \textsc{Subgroup-Level} data addition, this time from a distributional perspective. Using the subgroup similarity score introduced in Equation \ref{eq:subgroup-score-xy}, we find no statistically significant correlation between the distributional similarity of an added source with the test source and the added source's effect on subgroup test accuracy. This relationship, or lack thereof, is presented in Figure \ref{fig:subgroup-score}.

\section{Mean Consistency: A Barrier to Effective Data Intervention}
Following the findings in Section \ref{sec:naive-data-addition}, we shift our focus away from demographic differences between datasets, and instead look at the base rates themselves. One drawback of enforcing existing notions of fairness is that they lead to theoretically-bound tradeoffs with the overall performance of the model~\citep{10.1145/3097983.3098095, pmlr-v81-menon18a}. The incompatibility arises due to differing base rates between subgroups~\citep{DBLP:journals/corr/KleinbergMR16, zhao2022inherent, zliobaite}. This line of work implies that base rates should be examined when understanding the disparities induced by data. In our setting, we have the opportunity to favorably adjust subgroup base rates through data addition.

More  formally, we look at mean consistency, defined as the difference between a model's predictive mean and the true mean of the test set. Concretely, we define the \textit{mean discrepancy} between a learned predictor and a test dataset as follows:
\begin{equation}
    \textsc{Disc}(f, D) = \frac{1}{|D|} \bigg| \sum_{(x, y) \in D} \mathbf{1}[f(x) \ge t] - y \bigg|
    \label{eq:mean-discrepancy}
\end{equation}
This expresses the difference between the average output label of a model $f$ over a dataset $D$ and the true average value of the labels in $D$. 

Using this measure, we hypothesize that large mean discrepancies between subgroups are one source of unfairness in our datasets of interest. This can be understood as a situation where one subgroup exhibits starkly different outcome rates compared to others, leading to an overall training dataset and subsequent model mean which are skewed from the subgroup label mean. 

The formal relationship between mean discrepancy (Equation \ref{eq:mean-discrepancy}) and accuracy (Equation \ref{eq:acc}) can be derived using the triangle inequality:

\begin{align*}
    \textsc{Disc}(f, D) &\le \frac{1}{|D|}\sum_{(x, y \in D)} \bigg| \mathbf{1}[f(x) \ge t] - y \bigg| \\&\le 1 - \text{Acc}(f, D)
    \label{eq:mean-disc-acc}
\end{align*}

\begin{figure*}[t]
    \centering
    \includegraphics[width=1\linewidth]{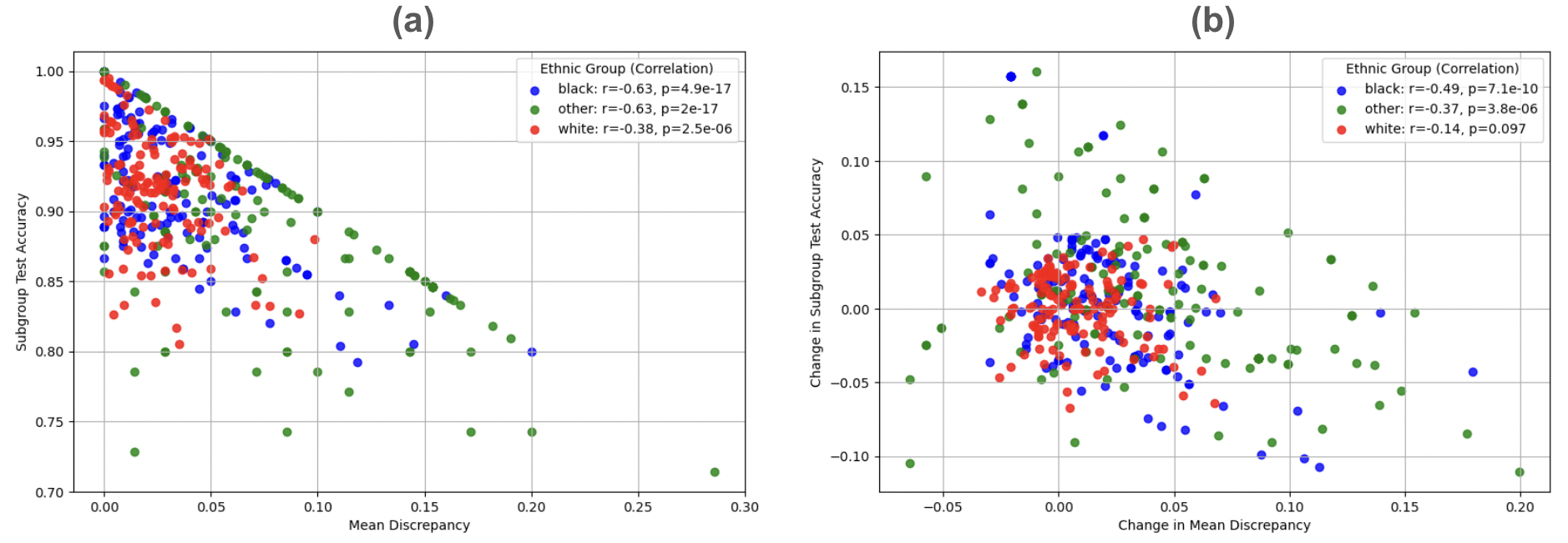}
    \caption{\small Mean consistency and subgroup performance: (a) Subgroup Mean Discrepancy (Eq. \ref{eq:mean-discrepancy}) vs. Subgroup Test Accuracy (Eq. \ref{eq:acc}). (b) Change in Subgroup Mean Discrepancy vs. Change in Subgroup Test Accuracy. Strong negative correlations are observed across all subgroups in (a), and in all minority subgroups in (b).}
    \label{fig:mean-discrepancy}
\end{figure*}

The mean discrepancy bounds the test accuracy, such that the greater the mean discrepancy of a model is, the lower its performance ceiling will be. This guarantee is empirically confirmed in Figure \ref{fig:mean-discrepancy}(a), where we exhibit a clean boundary line $y = 1-x$. We find that this correlation transfers to the data addition scenario---in Figure \ref{fig:mean-discrepancy}(b), we plot a statistically significant correlation between the change in subgroup mean discrepancy after adding a data source with the resultant change in subgroup accuracy. This trend is seen across all ethnic subgroups. This finding provides strong evidence that mean discrepancy plays a key role in determining the effect of data addition on subgroup performance.

The resultant data selection strategy is thus to add sources that correct the subgroup mean discrepancy exhibited over the base model. This leads to the following selection criteria:

\begin{equation*}
    i_* = \arg \min_{i} \textsc{Disc}(f_{D_{\text{train}+i}}, D_{\text{test}}^a) - \textsc{Disc}(f_{D_{\text{train}}}, D_{\text{test}}^a)
\end{equation*}

\begin{figure}
    \centering
    \includegraphics[width=0.8\linewidth]{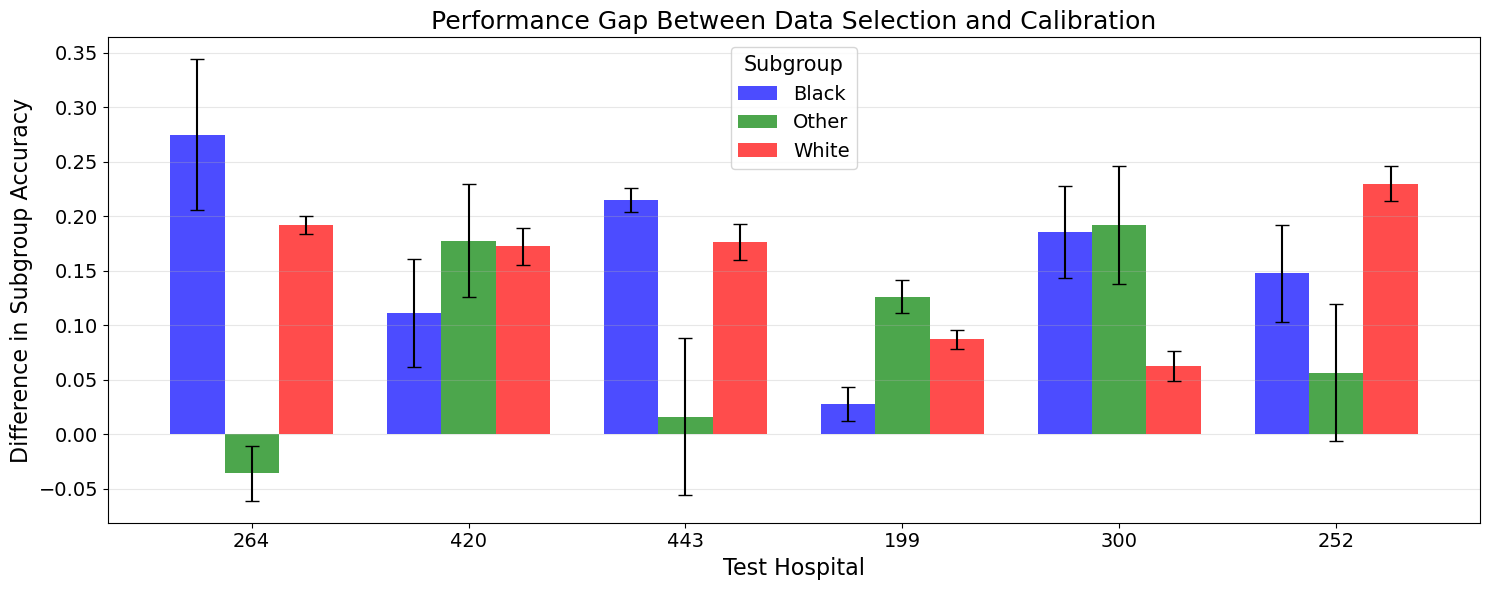}
    \caption{\small Difference in best-case subgroup performance (without calibration) and worst-case subgroup performance (with calibration) after \textsc{Whole-Source} data addition. The overwhelming majority of subgroups show positive differences, supporting the idea that making informed choices for data addition is more important than calibrating post-hoc.}
    \label{fig:calibration-v-addition}
\end{figure}

\paragraph{Data-Centric vs. Post-Processing Interventions}
Our empirical results complement existing techniques that use calibration to achieve better subgroup performance. Next, we apply post-processing calibration via Isotonic Regression on models trained using \textsc{Whole-Source} data addition. In these experiments, we additionally withhold a validation set of $200$ samples drawn from the same distribution (or mixture of distributions) as the training set. We use this data to perform calibration after training. 

\begin{table}[t]
\small
\centering
\setlength{\tabcolsep}{6pt}
\renewcommand{\arraystretch}{1.15}
\begin{tabular}{lcccccc}
\toprule
& \multicolumn{2}{c}{\textbf{Black}}
& \multicolumn{2}{c}{\textbf{Other}}
& \multicolumn{2}{c}{\textbf{White}} \\
\cmidrule(lr){2-3}\cmidrule(lr){4-5}\cmidrule(lr){6-7}
& W/o Cal. & With Cal.
& W/o Cal. & With Cal.
& W/o Cal. & With Cal. \\
\midrule
Best
& \textbf{4.2\%} $\pm$ 1.6 & 3.4\% $\pm$ 0.9
& 6.1\% $\pm$ 2.2 & \textbf{8.7\%} $\pm$ 1.3
& 2.3\% $\pm$ 0.3 & \textbf{2.8\%} $\pm$ 0.1 \\
Median
& 0.6\% $\pm$ 2.6 & 1.5\% $\pm$ 1.1
& 1.2\% $\pm$ 3.2 & 0.1\% $\pm$ 1.7
& 0.7\% $\pm$ 0.9 & 2.2\% $\pm$ 0.1 \\
Worst
& -4.1\% $\pm$ 2.8 & -5.9\% $\pm$ 4.3
& -3.4\% $\pm$ 2.8 & -13.0\% $\pm$ 4.4
& -3.4\% $\pm$ 2.7 & 1.2\% $\pm$ 0.4 \\
\bottomrule
\end{tabular}

\caption{Average change in subgroup accuracy for the Best, Median, and Worst \textsc{Whole-Source} data additions across all sources, with and without calibration. }
\label{tab:calibration-and-addition}
\end{table}

In Figure \ref{fig:calibration-v-addition}, we compare the change in subgroup accuracy after optimal data selection against calibration on sub-optimal data selection. We find that post-processing calibration is outperformed simply by choosing a better (or the best) available data source. 

However, we find that the combination of both data selection and calibration is generally most effective at improving subgroup performance. Table \ref{tab:calibration-and-addition} summarizes the average change in subgroup accuracy for the best, median, and worst available data sources with and without calibration. We find that the combination of data addition with calibration, on average, yields the highest best-case outcomes, along with better worst-case outcomes. 

Overall, we find that without intentional dataset selection that takes into account desired fairness outcomes, calibration alone does not guarantee improvement. Under this problem set-up where we have fine-grained control over dataset composition, we find it necessary to first make informed data addition choices in order to maximize the performance boost from post-processing calibration.

\section{Discussion} 
In this work, we investigate the landscape of data interventions for improving subgroup performance where multiple additional sources of data are available. Through comprehensive experiments across the eICU and MIMIC-IV datasets, we demonstrate that naive strategies for data addition, such as increasing subgroup representation or minimizing distribution shift, fail to reliably improve subgroup performance. We identify mean discrepancy as a key bottleneck of subgroup performance, and formalize mean consistency as a necessary criterion for successful data interventions.

We compare data interventions with post-processing calibration (model interventions), and find that ``fixing the data" is more important for improving subgroup performance than direct post-hoc calibration. When we combine both data selection and calibration, we observe the best possible performance across most subgroups. 


\subsection{Limitations \& Future Work}
Our experiments are purposefully narrowed to focus on ICU data using race as the sensitive attribute. This focus helps better isolate the the effects of data interventions on subgroup performance for a commonly task in clinical machine learning. Future works should investigate more tasks in the healthcare domain. Our ultimate hope is for a task-agnostic framework for the selection of data addition heuristics.

Our work presents promising findings for deeper investigations into task-specific data selection heuristics. Making informed choices about training data is one of the most consequential steps of the machine learning pipeline. We hope our findings motivate deeper investigations into the practical data decision-making process, and promote methods which are conscious of the unequal effects they can have on under-served and under-represented populations.

\bibliography{iclr2026_conference}
\bibliographystyle{iclr2026_conference}

\appendix
\onecolumn
\section{Datasets}

The demographic data for both datasets used in this study---eICU and MIMIC-IV--- are presented in this section. For both datasets, the examined sensitive attribute was patient self-reported race. We also provide information on the data sources used to stratify each dataset.

\subsection{eICU Collaborative Research Database}
Table \ref{tab:eicu-data} summarizes the subset of the eICU Collaborative Research Database used in this study. We treat different hospitals, denoted by a unique Hospital ID, as independent data sources. The full database includes 200,859 patient unit encounters in 208 hospitals across the US. For this study, we consider only the hospitals with over 2000 recorded patient encounters, a total of 12 data sources.

\begin{table}[h]
  \centering 
  \caption{Subgroup information across hospitals with $2000$+ patient encounters in the eICU Dataset}
    \begin{tabular}{llllll}
    \toprule
    \multicolumn{1}{c}{} & \multicolumn{5}{c}{\textbf{Count (Rate)}} \\
    \cmidrule(lr){2-6}  
    \textbf{Hos ID} & \textbf{Total}  & \textbf{Asian} & \textbf{Black} & \textbf{Other} & \textbf{White}  \\
    \midrule
    73 & 4320 & 61 (0.01) & 622 (0.15) & 347 (0.08) & 3221 (0.76) \\
    167 & 2107 & 29 (0.01) & 154 (0.07) & 421 (0.20) & 1503 (0.71) \\
    188 & 2299 & 29 (0.01) & 517 (0.22) & 64 (0.03) & 1689 (0.73) \\
    199 & 2529 & 3 (0.001) & 42 (0.02) & 48 (0.02) & 2434 (0.96) \\
    243 & 2812 & 24 (0.009) & 873 (0.31) & 83 (0.03) & 1831 (0.65) \\
    252 & 2210 & 7 (0.003) & 152 (0.07) & 50 (0.02) & 1993 (0.91) \\
    264 & 3745 & 31 (0.009) & 263 (0.07) & 64 (0.02) & 3299 (0.90) \\
    300 & 2370 & 19 (0.008) & 267 (0.11) & 84 (0.04) & 2000 (0.84) \\
    338 & 2762 & 5 (0.002) & 41 (0.01) & 143 (0.05) & 2568 (0.93) \\
    420 & 3425 & 52 (0.02) & 157 (0.05) & 276 (0.08) & 2940 (0.86) \\
    443 & 2580 & 12 (0.005) & 1352 (0.53) & 83 (0.03) & 1119 (0.44) \\
    458 & 2456 & 34 (0.01) & 747 (0.30) & 132 (0.05) & 1542 (0.63) \\
    \bottomrule
    \end{tabular}
  \label{tab:eicu-data} 
\end{table}

\subsection{MIMIC-IV}

The Medical Information Mart for Intensive Care (MIMIC)-IV contains data for over 65,000 patients admitted to an ICU and over 200,000 patients admitted to the emergency department between 2008-2022 at the Beth Israel Deaconess Medical Center in Boston, MA.

In this dataset, we treat the patient's admission department as their data source. All patient encounters are categorized into three admission types: Emergency, Elective, and Urgent. Patient circumstances are vastly different between the three departments, introducing sufficient distributional changes to justify their treatment as independent sources.

Table \ref{tab:mimic-data} summarizes the demographic rates across sources in the dataset.

\begin{table}[h]
  \centering 
  \caption{Subgroup information across admission types in the MIMIC-IV Dataset}
    \begin{tabular}{lllllll}
    \toprule
    \multicolumn{1}{c}{} & \multicolumn{5}{c}{\textbf{Count (Rate)}} \\
    \cmidrule(lr){2-7}  
    \textbf{Adm Type} & \textbf{Total}  & \textbf{Asian} & \textbf{Black} & \textbf{Hispanic} & \textbf{Other} & \textbf{White}  \\
    \midrule
    Elective & 3464 & 69 (0.02) & 155 (0.045) & 75 (0.021) & 478 (0.138) & 2687 (0.775) \\
    Emergency & 21669 & 502 (0.023) & 2155 (0.099) & 741 (0.034) & 2965 (0.137) & 15306 (0.706) \\
    Urgent & 746 & 12 (0.016) & 17 (0.022) & 16 (0.021) & 318 (0.42) & 383 (0.51) \\
    \bottomrule
    \end{tabular}
  \label{tab:mimic-data} 
\end{table}

\section{Results from the MIMIC-IV Dataset}
\subsection{Experiment Details}
To confirm our findings in the eICU dataset, we conducted additional experiments on MIMIC-IV~\citep{johnson2023mimic}, a freely accessible electronic health record dataset. 

We predict binary in-hospital mortality using diagnosis codes, procedures, and lab measurements extracted from the patient record. In the database, there are $9$ possible admission types, including urgent care, surgical same-day, and emergency ambulatory services. We group the admission types into three broad categories to ensure sufficient sample sizes in each---Urgent, Elective, and Emergency care. We treat each of these admission categories as an independent data ``source". Subgroups are patient self-reported race.

\subsection{Results and Analysis}
The results of our data addition experiments are shown in Table \ref{tab:mimic-data-addition}. Performance changes larger than $0.5\%$ in either direction are denoted with a colored arrow.

For models trained on combined datasets, we find large inconsistencies in the performance changes across subgroups. For example, we find that the overall accuracy in Urgent care is notably improved when adding data from Elective departments, primarily as a result of significant improvements among the Asian, Other, and White populations. However, this improvement is also at the expense of the Black and Hispanic populations, who face $0.1\%$ and $1.1\%$ decreases in test accuracy, respectively. The decline in Hispanic performance is larger than improvements to any subgroup, begging the question of whether this addition would be desirable.

\begin{table}
  \centering 
  \caption{Change in overall \& subgroup performance after \textsc{Whole-Source} data addition.}
    \begin{tabular}{llllllll}
    \toprule
    \multicolumn{2}{c}{\textbf{Sources}} & \multicolumn{6}{c}{\textbf{Change in ACC ($\Delta$)}} \\
    \cmidrule(lr){1-2}\cmidrule(lr){3-8} 
    \textbf{Test} & \textbf{Added}  & \textbf{Asian} & \textbf{Black} & \textbf{Hispanic} & \textbf{Other} & \textbf{White} & \textbf{Overall} \\
    \midrule
    Elective & Elective & 0.001 & 0.006 \textcolor{green}{$\uparrow$} & 0.007 \textcolor{green}{$\uparrow$} & 0.006 \textcolor{green}{$\uparrow$} & -0.001 & 0.000 \\
    Elective & Emergency & 0.001 & -0.001 & 0.010 \textcolor{green}{$\uparrow$} & -0.003 & 0.001 & 0.001 \\
    Elective & Urgent & -0.001 & 0.000 & 0.003 & -0.003 & 0.002 & -0.001 \\
    Emergency & Elective & -0.001 & -0.005 \textcolor{red}{$\downarrow$} & -0.003 & 0.010 \textcolor{green}{$\uparrow$} & 0.000 & -0.001 \\
    Emergency & Emergency & 0.002 & -0.010 \textcolor{red}{$\downarrow$} & -0.004 & 0.020 \textcolor{green}{$\uparrow$} & 0.008 \textcolor{green}{$\uparrow$} & 0.001 \\
    Emergency & Urgent & -0.001 & -0.021 \textcolor{red}{$\downarrow$} & -0.001 & -0.003 & 0.001 & -0.001 \\
    Urgent & Elective & 0.007 \textcolor{green}{$\uparrow$} & -0.001 & -0.011 \textcolor{red}{$\downarrow$} & 0.007 \textcolor{green}{$\uparrow$} & 0.006 \textcolor{green}{$\uparrow$} & 0.009 \textcolor{green}{$\uparrow$} \\
    Urgent & Emergency & 0.005 \textcolor{green}{$\uparrow$} & -0.042 \textcolor{red}{$\downarrow$} & -0.018 \textcolor{red}{$\downarrow$} & 0.007 \textcolor{green}{$\uparrow$} & 0.016 \textcolor{green}{$\uparrow$} & -0.001 \\
    Urgent & Urgent & 0.002 & -0.012 \textcolor{red}{$\downarrow$} & 0.018 \textcolor{green}{$\uparrow$} & 0.000 & 0.001 & 0.004 \\
    \bottomrule
    \end{tabular}
    
  \label{tab:mimic-data-addition} 
\end{table}

\paragraph{Evaluation of Existing Data Addition Methods}
Tables \ref{tab:mimic-subgroup-rate} and \ref{tab:mimic-samples-added} look at existing data addition strategies under the lens of fairness. They corroborate the results seen on the eICU dataset in Figures \ref{fig:subgroup-rate} and \ref{fig:samples-added}, respectively, showing that these data balancing strategies are ineffective at explaining the performance gaps between subgroups.


\begin{table}[tp]
\centering
\caption{Change in subgroup ratio and resulting change in subgroup accuracy after \textsc{Whole-Source} data addition on the MIMIC-IV dataset.}
\label{tab:mimic-subgroup-rate}
\vspace{10px}

\begin{tabular}{lllc>{\centering\arraybackslash}p{1.5cm}c}
\toprule
\textbf{Ethnicity} & \textbf{Admission Type} & \textbf{Added Source} & \textbf{Rate Change} & \textbf{Acc. Diff} & \textbf{Correlation} \\
\midrule
\multirow[=]{9}{*}{Asian} 
  & Elective  & Elective  & 0.000  & 0.006 & \multirow[=]{9}{*}{\makecell{$r=-0.28$,\\ $p=0.47$}} \\
  & Elective  & Emergency  & 0.002  & -0.001 & \\
  & Elective  & Urgent  & -0.002  & 0.000 & \\
  & Emergency  & Elective  & -0.002  & -0.005 & \\
  & Emergency  & Emergency  & 0.000  & -0.010 & \\
  & Emergency  & Urgent  & -0.004  & -0.021 & \\
  & Urgent  & Elective  & 0.002  & -0.001 & \\
  & Urgent  & Emergency  & 0.004  & -0.042 & \\
  & Urgent  & Urgent  & 0.000  & -0.012 & \\
\midrule
\multirow[=]{9}{*}{Black} 
  & Elective  & Elective  & 0.000  & 0.007 & \multirow{9}{*}{\makecell{$r=-0.20$,\\ $p=0.61$}} \\
  & Elective  & Emergency  & 0.027  & 0.010 & \\
  & Elective  & Urgent  & -0.011  & 0.003 & \\
  & Emergency  & Elective  & -0.027  & -0.003 & \\
  & Emergency  & Emergency  & 0.000  & -0.004 & \\
  & Emergency  & Urgent  & -0.038  & -0.001 & \\
  & Urgent  & Elective  & 0.011  & -0.011 & \\
  & Urgent  & Emergency  & 0.038  & -0.018 & \\
  & Urgent  & Urgent  & 0.000  & 0.018 & \\
\midrule
\multirow[=]{9}{*}{Hispanic} 
  & Elective  & Elective  & 0.000  & 0.006 & \multirow[=]{9}{*}{\makecell{$r=-0.05$,\\ $p=0.89$}} \\
  & Elective  & Emergency  & 0.006  & -0.003 & \\
  & Elective  & Urgent  & -0.000  & -0.003 & \\
  & Emergency  & Elective  & -0.006  & 0.010 & \\
  & Emergency  & Emergency  & 0.000  & 0.020 & \\
  & Emergency  & Urgent  & -0.006  & -0.003 & \\
  & Urgent  & Elective  & 0.000  & 0.007 & \\
  & Urgent  & Emergency  & 0.006  & 0.007 & \\
  & Urgent  & Urgent  & 0.000  & 0.000 & \\
\midrule
\multirow[=]{9}{*}{Other} 
  & Elective  & Elective  & 0.000  & -0.001 & \multirow[=]{9}{*}{\makecell{$r=-0.62$,\\ $p=0.08$}} \\
  & Elective  & Emergency  & -0.001  & 0.001 & \\
  & Elective  & Urgent  & 0.144  & 0.002 & \\
  & Emergency  & Elective  & 0.001  & 0.000 & \\
  & Emergency  & Emergency  & 0.000  & 0.008 & \\
  & Emergency  & Urgent  & 0.145  & 0.001 & \\
  & Urgent  & Elective  & -0.144  & 0.006 & \\
  & Urgent  & Emergency  & -0.145  & 0.016 & \\
  & Urgent  & Urgent  & 0.000  & 0.001 & \\
\midrule
\multirow[=]{9}{*}{White} 
  & Elective  & Elective  & 0.000  & 0.000 & \multirow[=]{9}{*}{\makecell{$r=0.58$,\\ $p=0.10$}} \\
  & Elective  & Emergency  & -0.035  & 0.001 & \\
  & Elective  & Urgent  & -0.131  & -0.001 & \\
  & Emergency  & Elective  & 0.035  & -0.001 & \\
  & Emergency  & Emergency  & 0.000  & 0.001 & \\
  & Emergency  & Urgent  & -0.096  & -0.001 & \\
  & Urgent  & Elective  & 0.131  & 0.009 & \\
  & Urgent  & Emergency  & 0.096  & -0.001 & \\
  & Urgent  & Urgent  & 0.000  & 0.004 & \\
\end{tabular}
\end{table}

\begin{table}[tp]
\centering
\caption{Subgroup samples added and resulting change in subgroup accuracy after \textsc{Subgroup-Only} data addition on the MIMIC-IV dataset.}
\label{tab:mimic-samples-added}
\vspace{10px}

\begin{tabular}{lllc>{\centering\arraybackslash}p{1.5cm}c}
\toprule
\textbf{Ethnicity} & \textbf{Admission Type} & \textbf{Added Source} & \textbf{Samples Added} & \textbf{Acc. Diff} & \textbf{Correlation} \\
\midrule
\multirow[=]{9}{*}{Asian} 
  & Elective  & Elective  & 19  & 0.006 & \multirow[=]{9}{*}{\makecell{$r=-0.24$,\\ $p=0.54$}} \\
  & Elective  & Emergency  & 23  & -0.001 & \\
  & Elective  & Urgent  & 16  & 0.000 & \\
  & Emergency  & Elective  & 19  & -0.005 & \\
  & Emergency  & Emergency  & 23  & -0.010 & \\
  & Emergency  & Urgent  & 16  & -0.021 & \\
  & Urgent  & Elective  & 19  & -0.001 & \\
  & Urgent  & Emergency  & 23  & -0.042 & \\
  & Urgent  & Urgent  & 16  & -0.012 & \\
\midrule
\multirow[=]{9}{*}{Black} 
  & Elective  & Elective  & 44  & 0.007 & \multirow[=]{9}{*}{\makecell{$r=-0.37$,\\ $p=0.33$}} \\
  & Elective  & Emergency  & 99  & 0.010 & \\
  & Elective  & Urgent  & 22  & 0.003 & \\
  & Emergency  & Elective  & 44  & -0.003 & \\
  & Emergency  & Emergency  & 99  & -0.004 & \\
  & Emergency  & Urgent  & 22  & -0.001 & \\
  & Urgent  & Elective  & 44  & -0.011 & \\
  & Urgent  & Emergency  & 99  & -0.018 & \\
  & Urgent  & Urgent  & 22  & 0.018 & \\
\midrule
\multirow[=]{9}{*}{Hispanic} 
  & Elective  & Elective  & 21  & 0.006 & \multirow[=]{9}{*}{\makecell{$r=0.33$,\\ $p=0.38$}} \\
  & Elective  & Emergency  & 34  & -0.003 & \\
  & Elective  & Urgent  & 21  & -0.003 & \\
  & Emergency  & Elective  & 21  & 0.010 & \\
  & Emergency  & Emergency  & 34  & 0.020 & \\
  & Emergency  & Urgent  & 21  & -0.003 & \\
  & Urgent  & Elective  & 21  & 0.007 & \\
  & Urgent  & Emergency  & 34  & 0.007 & \\
  & Urgent  & Urgent  & 21  & 0.000 & \\
\midrule
\multirow[=]{9}{*}{Other} 
  & Elective  & Elective  & 137  & -0.001 & \multirow[=]{9}{*}{\makecell{$r=-0.36$,\\ $p=0.34$}} \\
  & Elective  & Emergency  & 136  & 0.001 & \\
  & Elective  & Urgent  & 426  & 0.002 & \\
  & Emergency  & Elective  & 137  & 0.000 & \\
  & Emergency  & Emergency  & 136  & 0.008 & \\
  & Emergency  & Urgent  & 426  & 0.001 & \\
  & Urgent  & Elective  & 137  & 0.006 & \\
  & Urgent  & Emergency  & 136  & 0.016 & \\
  & Urgent  & Urgent  & 426  & 0.001 & \\
\midrule
\multirow[=]{9}{*}{White} 
  & Elective  & Elective  & 775  & 0.000 & \multirow[=]{9}{*}{\makecell{$r=0.23$,\\ $p=0.55$}} \\
  & Elective  & Emergency  & 706  & 0.001 & \\
  & Elective  & Urgent  & 513  & -0.001 & \\
  & Emergency  & Elective  & 775  & -0.001 & \\
  & Emergency  & Emergency  & 706  & 0.001 & \\
  & Emergency  & Urgent  & 513  & -0.001 & \\
  & Urgent  & Elective  & 775  & 0.009 & \\
  & Urgent  & Emergency  & 706  & -0.001 & \\
  & Urgent  & Urgent  & 513  & 0.004 & \\
\end{tabular}
\end{table}

\section{Supplemental Figures}

In this section, we provide supplemental figures and results that were not included in the main paper.

\subsection{Full Results of \textsc{Whole-Source} Data Addition}

Figure \ref{fig:data_addition_heatmaps} shows the full set of results measuring the change in accuracy after \textsc{Whole-Source} data addition in the eICU dataset. The results are presented for overall as well as subgroup-level test accuracy.

\begin{figure}
    \centering
    \includegraphics[width=1\linewidth]{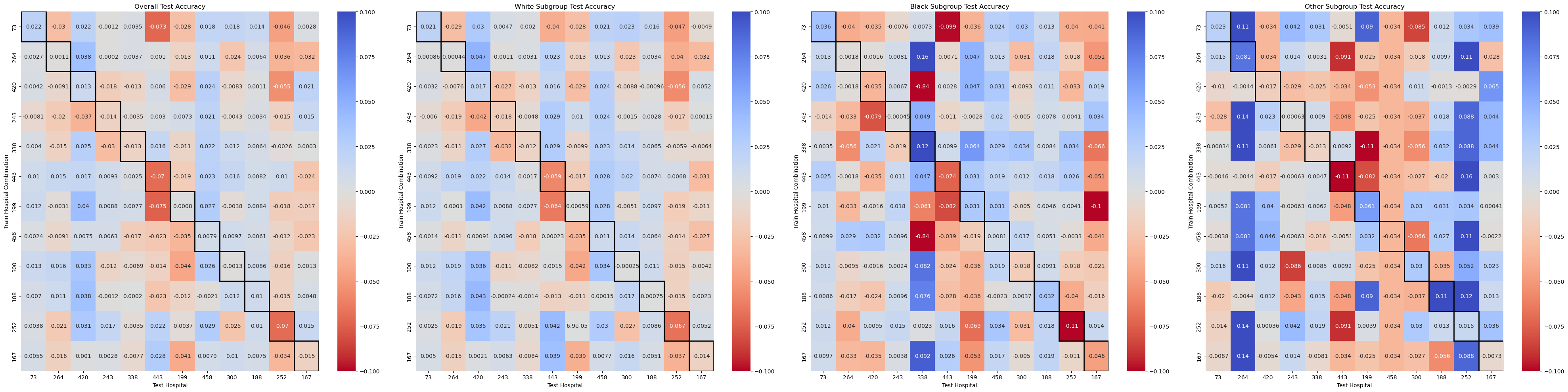}
    \caption{\small Change in Overall and subgroup-level accuracy after \textsc{Whole-Source} data addition. All results from abridged Figure \ref{fig:data_addition_heatmaps}.}
\end{figure}

\subsection{Pareto Frontier}

\begin{figure}[h]
    \centering
    \includegraphics[width=\linewidth]{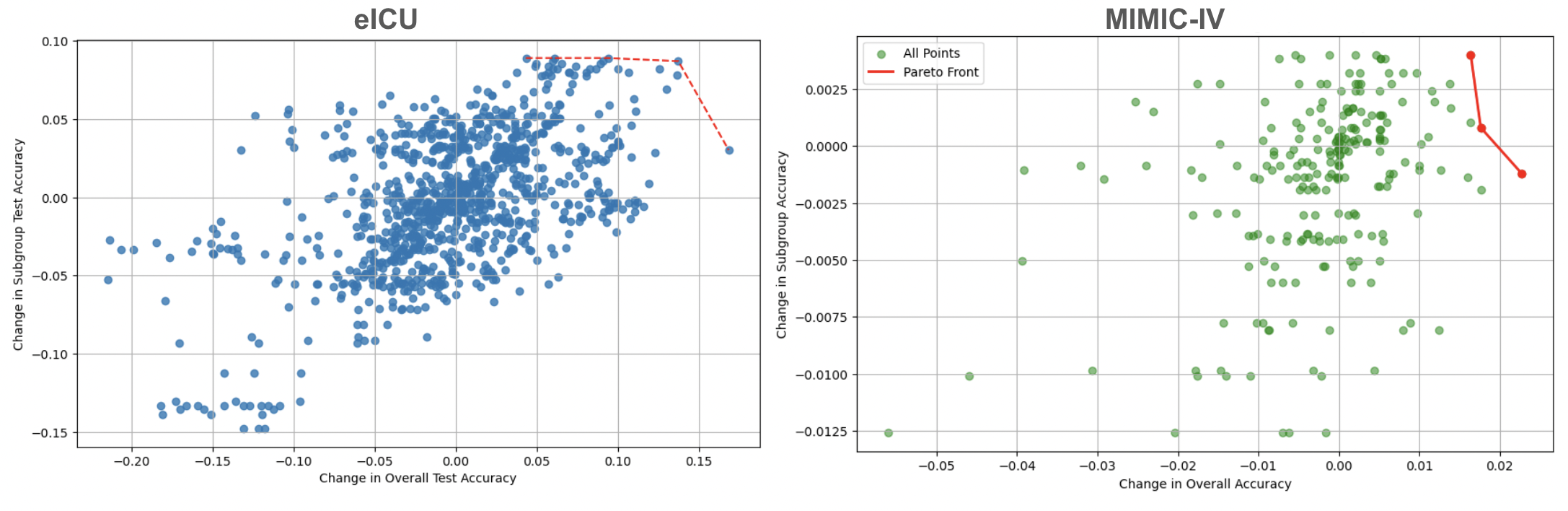}
    \caption{Change in overall test accuracy vs change in subgroup test accuracy. The Pareto Frontier is plotted on each subfigure to visualize the tradeoff between fairness and performance.}
    \label{fig:pareto}
\end{figure}

We want to evaluate whether our data-centric methods for fairness are able to bypass the commonly-studied tradeoff between subgroup performance and overall performance. In these figures, we plot the Pareto Frontier of \textsc{Whole-Source} data addition in both the eICU and MIMIC-IV datasets. We do not observe any considerable tradeoff between overall and subgroup performance, a promising result. These results are visualized in \ref{fig:pareto}.

\section{AUC Figures}

We strengthen the findings our eICU experiments by extending evaluation to additional performance metrics. This section reports the findings of data addition on overall and subgroup AUC, defined in \ref{eq:auc}. When sample sizes from minority groups are especially small, \citeauthor{zhioua2025originssamplingbiasimplications} recommend using metrics which consider the tradeoff between sensitivity and specificity, such as AUC. 

\subsection{Results and Analysis}

\begin{figure}[p]
    \centering
    \includegraphics[width=1\linewidth]{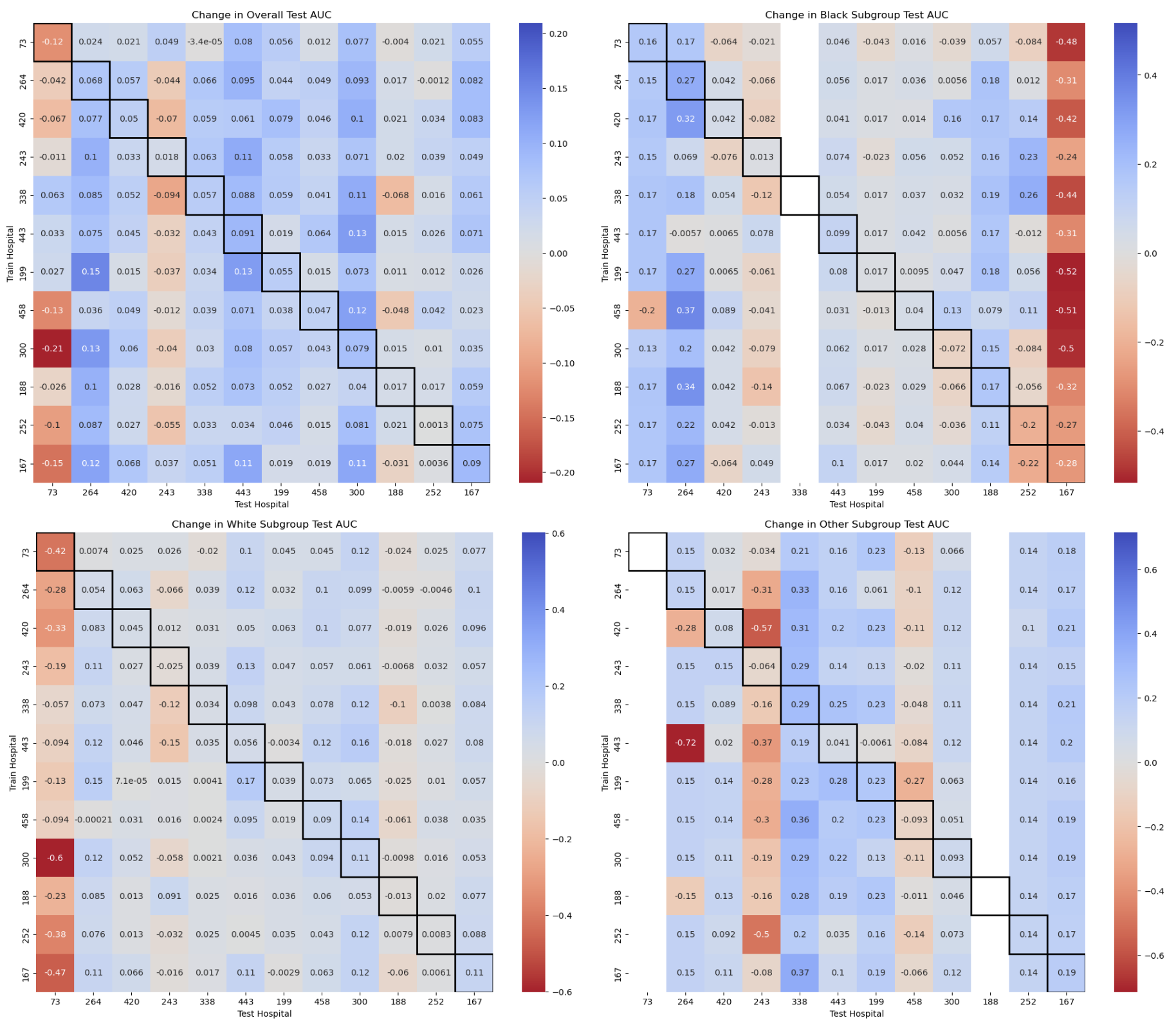}
    \caption{Change in Overall and Subgroup AUC by ethnicity group after \textsc{Whole-Source} data addition on the eICU dataset. Experiments are identical to the ones used to produce \ref{fig:data_addition_heatmaps}.}
    \label{fig:auc-heatmaps}
\end{figure}

Figure \ref{fig:auc-heatmaps} shows the change in AUC after data addition compared to the base results. Observing the plots, we see a similar phenomenon as Figure \ref{fig:data_addition_heatmaps} where the performance changes due to data addition are unequally distributed across subgroups. Most notably, Test Hospital $167$ is sees improvements in Overall AUC in all addition cases, whereas the Black subpopulation exhibits significant declines in performance in all cases.

In order to compute a meaningful AUC, both positive and negative samples must be included in the validation set. For splits where a subgroup did not have both classes represented, we leave the AUC values empty.

\begin{figure}
    \centering
    \includegraphics[width=0.7\linewidth]{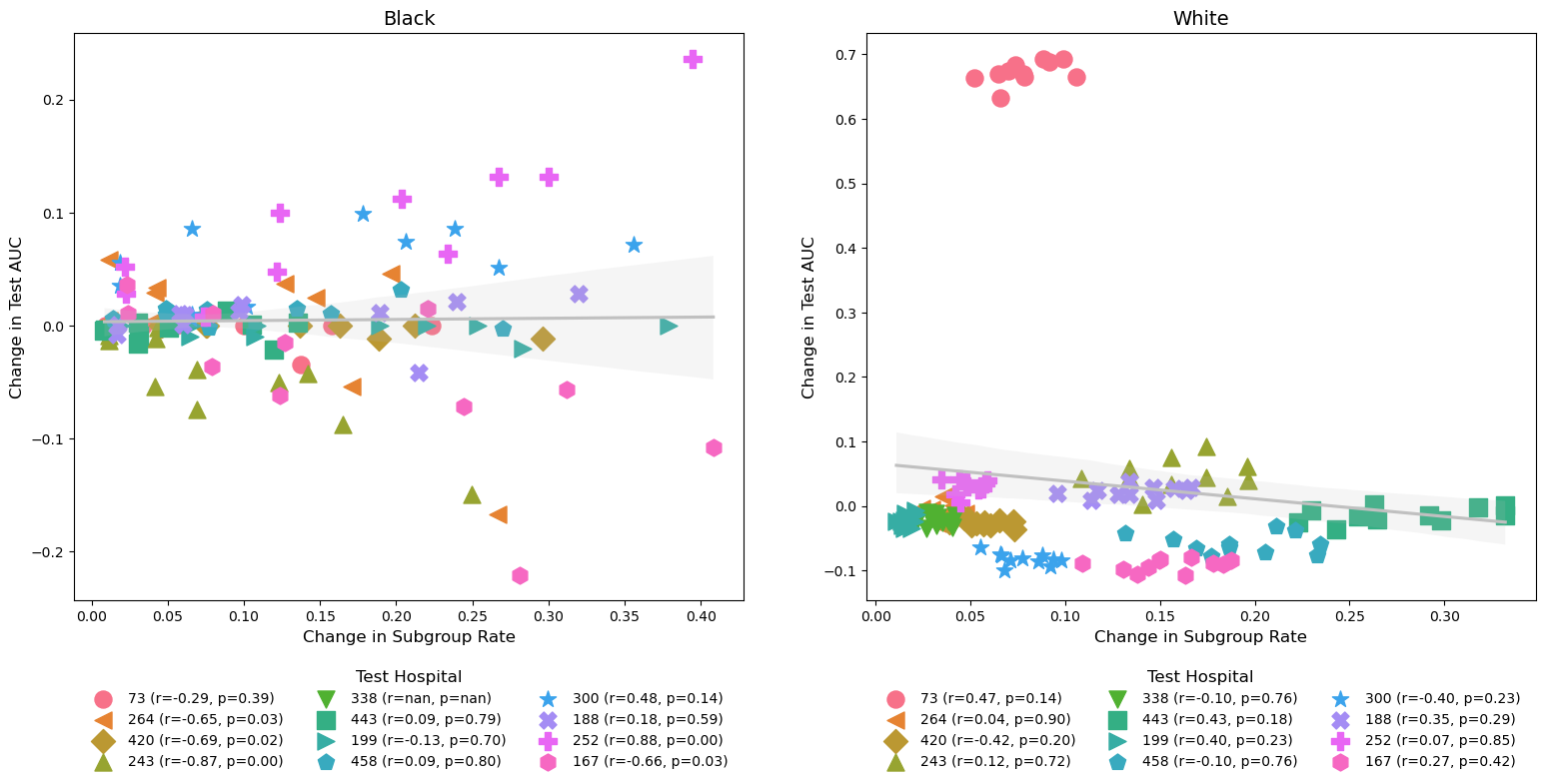}
    \caption{Change in subgroup ratio vs. Change in subgroup test AUC on the eICU dataset. Same experiment conditions as were used to produce \ref{fig:subgroup-rate}.}
    \label{fig:auc-subgroup-rate}
\end{figure}
\begin{figure}
    \centering
    \includegraphics[width=0.7\linewidth]{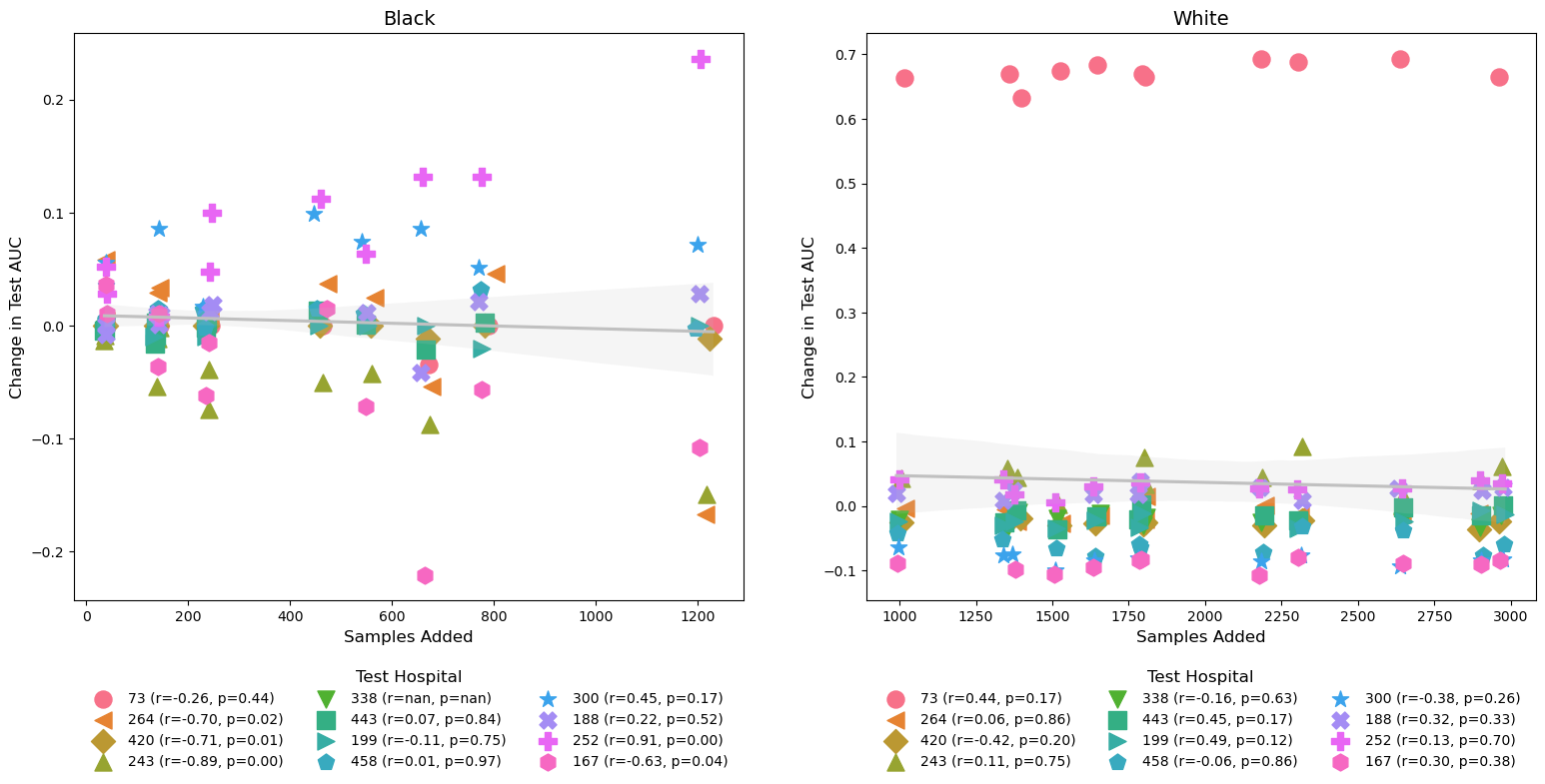}
    \caption{Number of subgroup samples added vs. Change in subgroup test AUC on the eICU dataset. Same experiment conditions as were used to produce \ref{fig:samples-added}.}
    \label{fig:auc-samples-added}
\end{figure}
\begin{figure}
    \centering
    \includegraphics[width=0.8\linewidth]{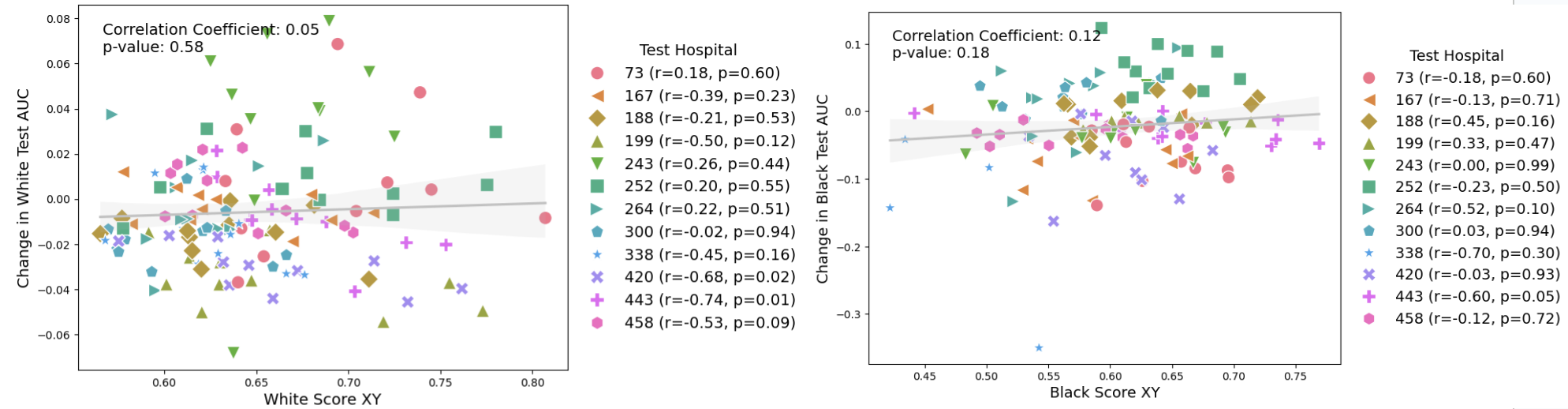}
    \caption{Subgroup similarity score vs. Change in subgroup test AUC on the eICU dataset. Same experiment conditions as were used to produce \ref{fig:subgroup-score}.}
    \label{fig:auc-subgroup-score}
\end{figure}

In Figures \ref{fig:auc-subgroup-rate}, \ref{fig:auc-samples-added}, and \ref{fig:auc-subgroup-score}, we find analogous results to Figures \ref{fig:subgroup-rate}, \ref{fig:samples-added}, and \ref{fig:subgroup-score}, respectively. In these experiments, we evaluate existing data interventions for fairness and find that they are unreliable in the scenario of OOD data addition. We do not find statistically significant relationships between the change in subgroup AUC with the change in subgroup samples, subgroup rate, or with the similarity of the added data source.

\begin{figure}
    \centering
    \includegraphics[width=1\linewidth]{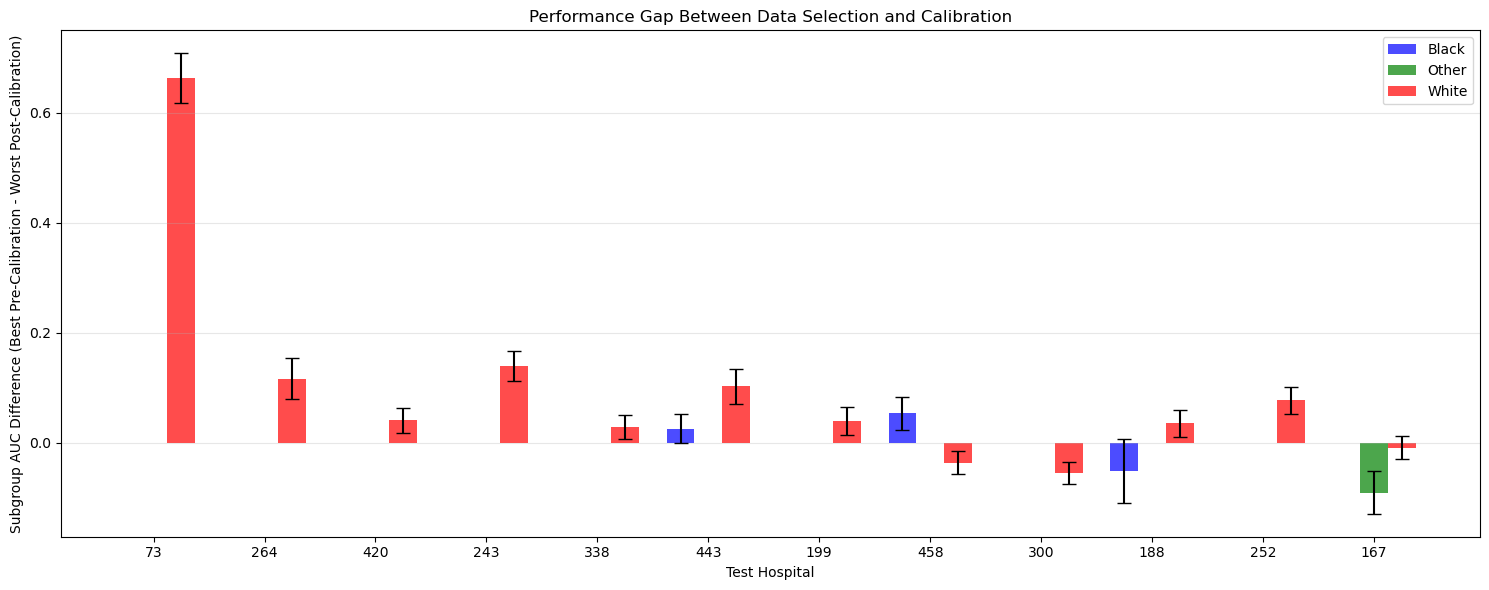}
    \caption{Difference in best-case subgroup AUC (without calibration) and worst-case subgroup AUC (with calibration) on the eICU dataset. Same experiment conditions as were used to produce \ref{fig:calibration-v-addition}.}
    \label{fig:auc-calibration-v-addition}
\end{figure}

\begin{figure}
    \centering
    \includegraphics[width=1\linewidth]{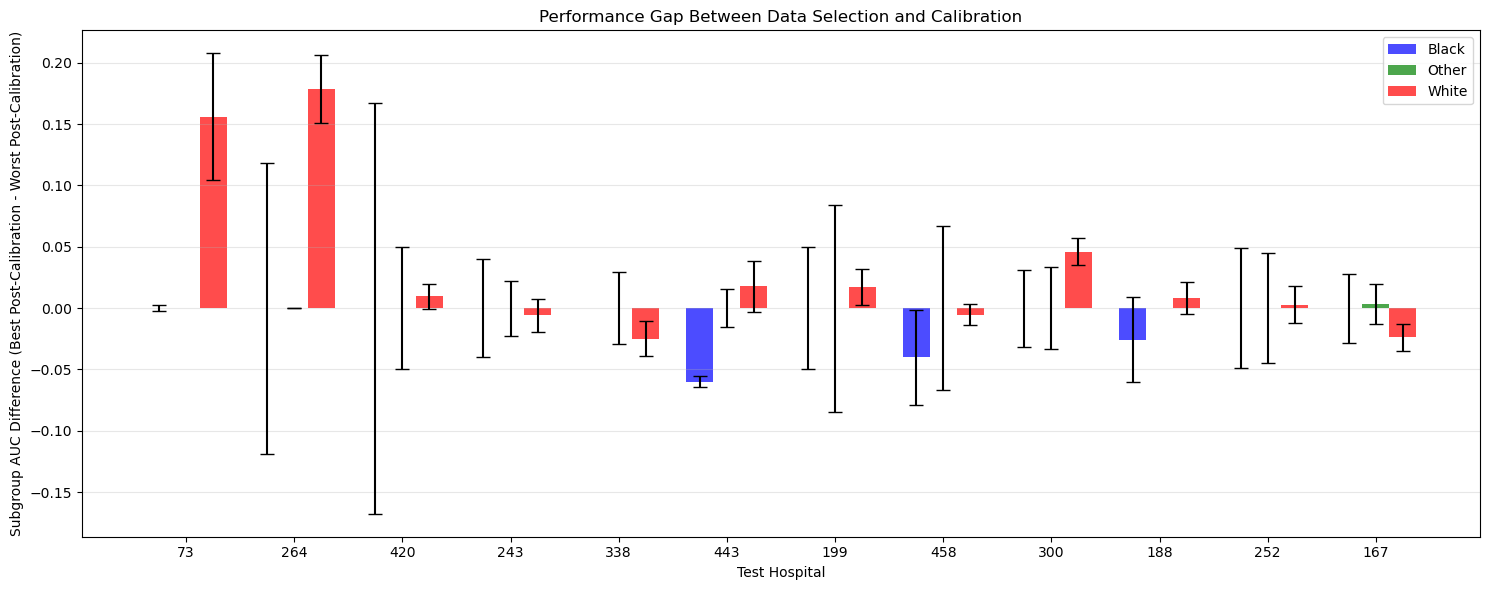}
    \caption{Difference in best-case subgroup AUC (with calibration) and worst-case subgroup AUC (with calibration) on the eICU dataset. Same experiment conditions as were used to produce \ref{tab:calibration-and-addition}.}
    \label{fig:auc-calibration-and-addition}
\end{figure}

Lastly, we reproduce Figures \ref{fig:calibration-v-addition} and \ref{tab:calibration-and-addition} looking at subgroup AUC, the results of which are shown in \ref{fig:auc-calibration-v-addition} and \ref{fig:auc-calibration-and-addition}, respectively. Although many samples are null, due to lack of positive samples in the validation set used for calibration, we still observe that performance gaps are generally positive. However, because the definition of calibration inherently includes accuracy, we do not observe as large of an effect on AUC as we did with accuracy.

\section{Extended Model Classes}
This section shares the results and visualizations for all experiments conducted using Light Gradient Boosting Machine (LGBM) and Long Short-Term Memory (LSTM) models. The figures in the main paper are all showing the results using a Logistic Regression (LR) model. Each experiment was replicated on each model class. In general, it is observed that all model classes exhibit similar behavior under data addition.

\subsection{Light Gradient Boosting Machine (LGBM) Results}

\begin{figure}
    \centering
    \includegraphics[width=1\linewidth]{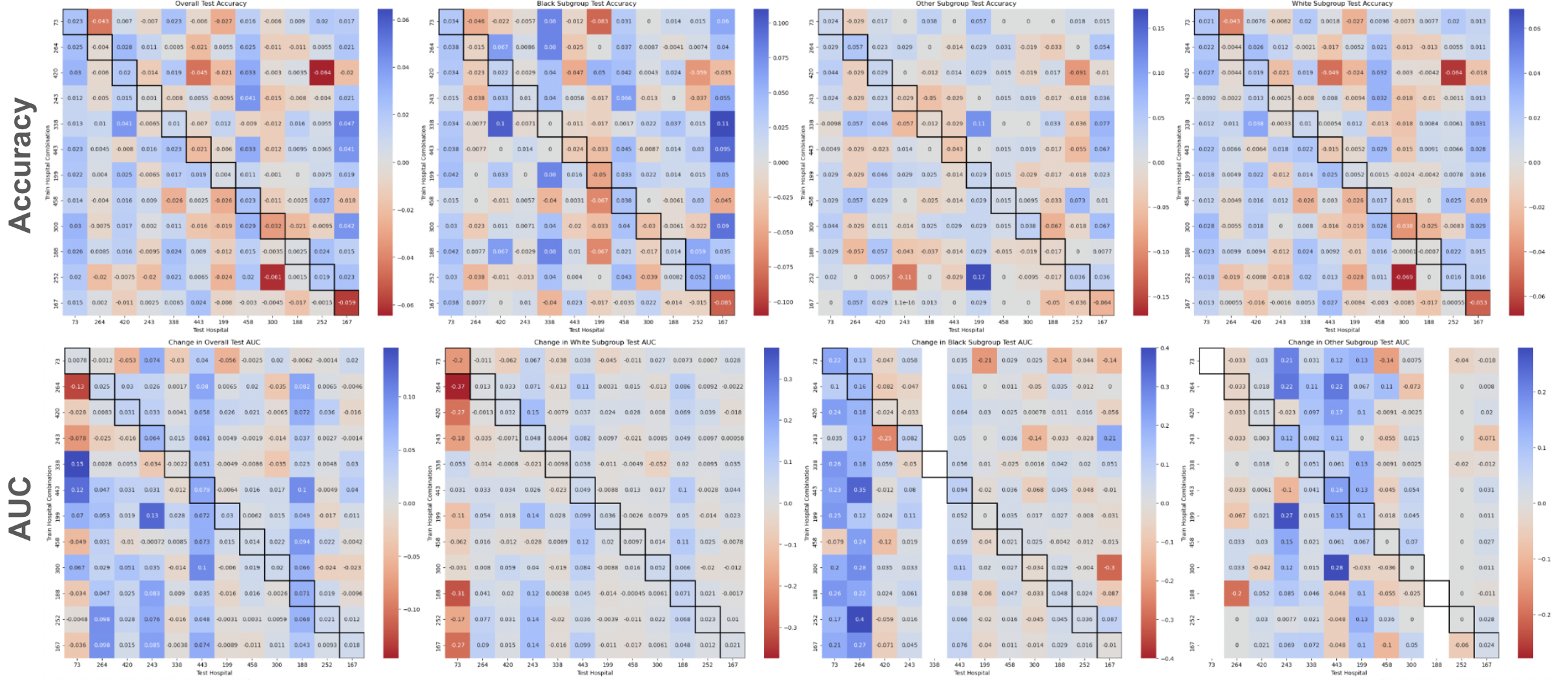}
    \caption{Change in performance after \textsc{Whole-Source} data addition using the Light Gradient Boosting Machine (LGBM) classifier on the eICU Dataset.}
    \label{fig:lgbm-heatmaps}
\end{figure}

The change in performance after \textsc{Whole-Source} data addition is shown in Figure \ref{fig:lgbm-heatmaps}. We visualize the data addition combinations for all 12 hospitals used in the eICU Dataset. Results are shown for both accuracy and AUC at both the overall and subgroup levels.

\begin{figure}
    \centering
    \includegraphics[width=0.5\linewidth]{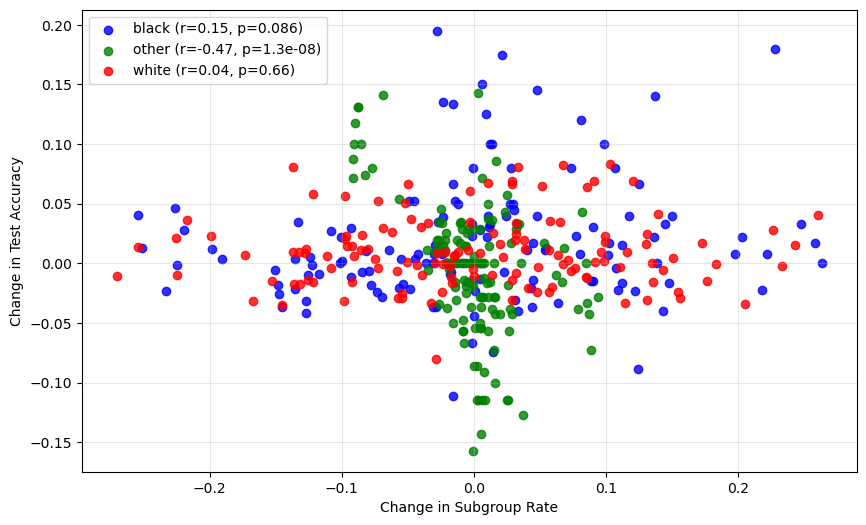}
    \caption{Change in Subgroup Rate vs. Change in Subgroup Accuracy after \textsc{Whole-Source} data addition, trained using an LGBM model.}
    \label{fig:lgbm-subgroup_rate}
\end{figure}

\begin{figure}
    \centering
    \includegraphics[width=1\linewidth]{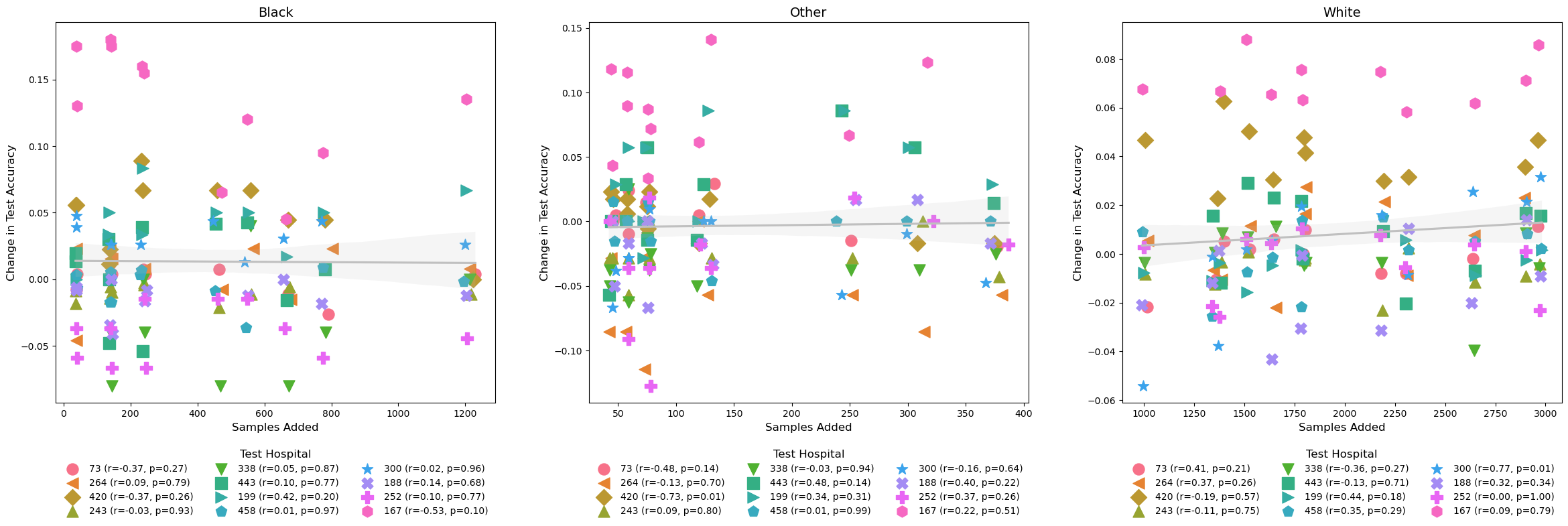}
    \caption{Number of subgroup samples added vs. Change in Subgroup Accuracy after \textsc{Subgroup-Level} data addition, trained using an LGBM model.}
    \label{fig:lgbm-samples_added}
\end{figure}

\begin{figure}
    \centering
    \includegraphics[width=0.9\linewidth]{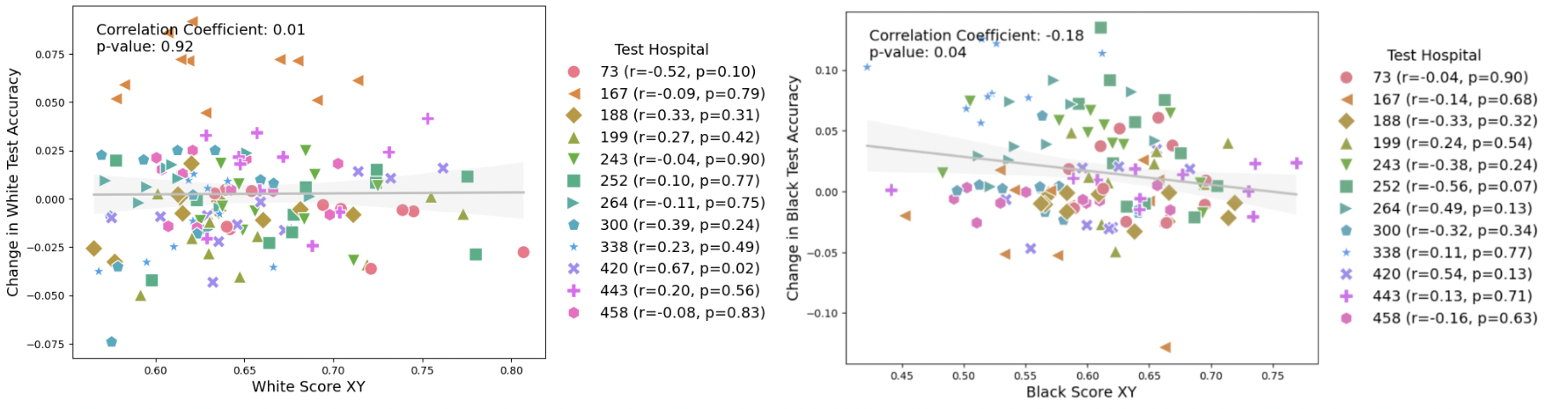}
    \caption{Subgroup Similarity Score between the test subgroup and added subgroup vs. Change in Subgroup Accuracy after \textsc{Subgroup-Level} data addition, trained using an LGBM model.}
    \label{fig:lgbm-subgroup_score}
\end{figure}

We also analyze the results of naive data selection strategies---maximizing subgroup rate, samples added, and similarity score---in Figures \ref{fig:lgbm-subgroup_rate}, \ref{fig:lgbm-samples_added}, and \ref{fig:lgbm-subgroup_score}, respectively. These are evaluated on the eICU dataset. We find that these existing data selection strategies are similarly ineffective when training an LGBM classifier.

\begin{figure}
    \centering
    \includegraphics[width=1\linewidth]{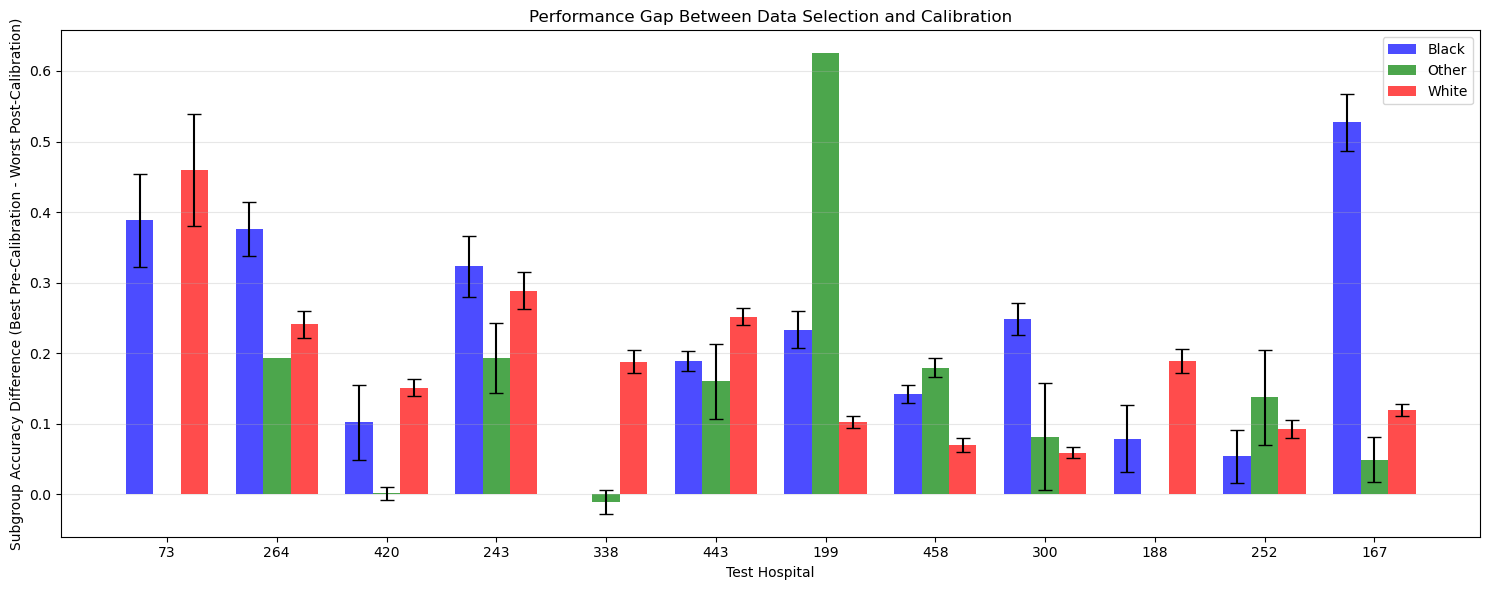}
    \caption{Difference in best-case subgroup performance (without calibration) and worst-case subgroup performance (with calibration) after \textsc{Whole-Source} data addition. Performance is evaluated on the eICU dataset using the subgroup accuracy metric and trained using an LGBM classifier.}
    \label{fig:lgbm-calibration-v-addition}
\end{figure}

\begin{figure}
    \centering
    \includegraphics[width=1\linewidth]{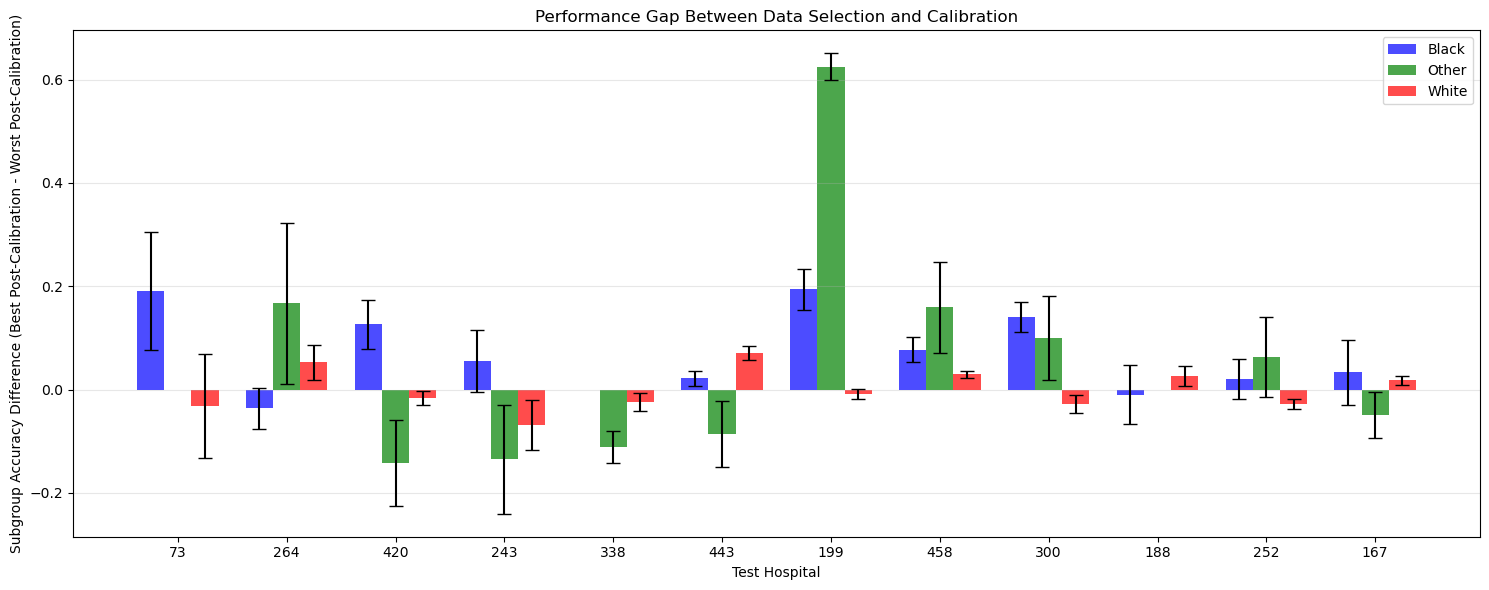}
    \caption{Difference in best-case subgroup performance (with calibration) and worst-case subgroup performance (with calibration) after \textsc{Whole-Source} data addition. Performance is evaluated on the eICU dataset using the subgroup accuracy metric and trained using an LGBM classifier.}
    \label{fig:lgbm-calibration-and-addition}
\end{figure}

We compare data addition and calibration in \ref{fig:lgbm-calibration-v-addition}, finding that, in all cases, informed data selection outperforms calibration as a fairness intervention strategy. Finally, in \ref{fig:lgbm-calibration-and-addition}, we compare the combination of data selection and calibration on an LGBM classifier. 

\subsection{LSTM Results}

\begin{figure}[p]
    \centering
    \includegraphics[width=1\linewidth]{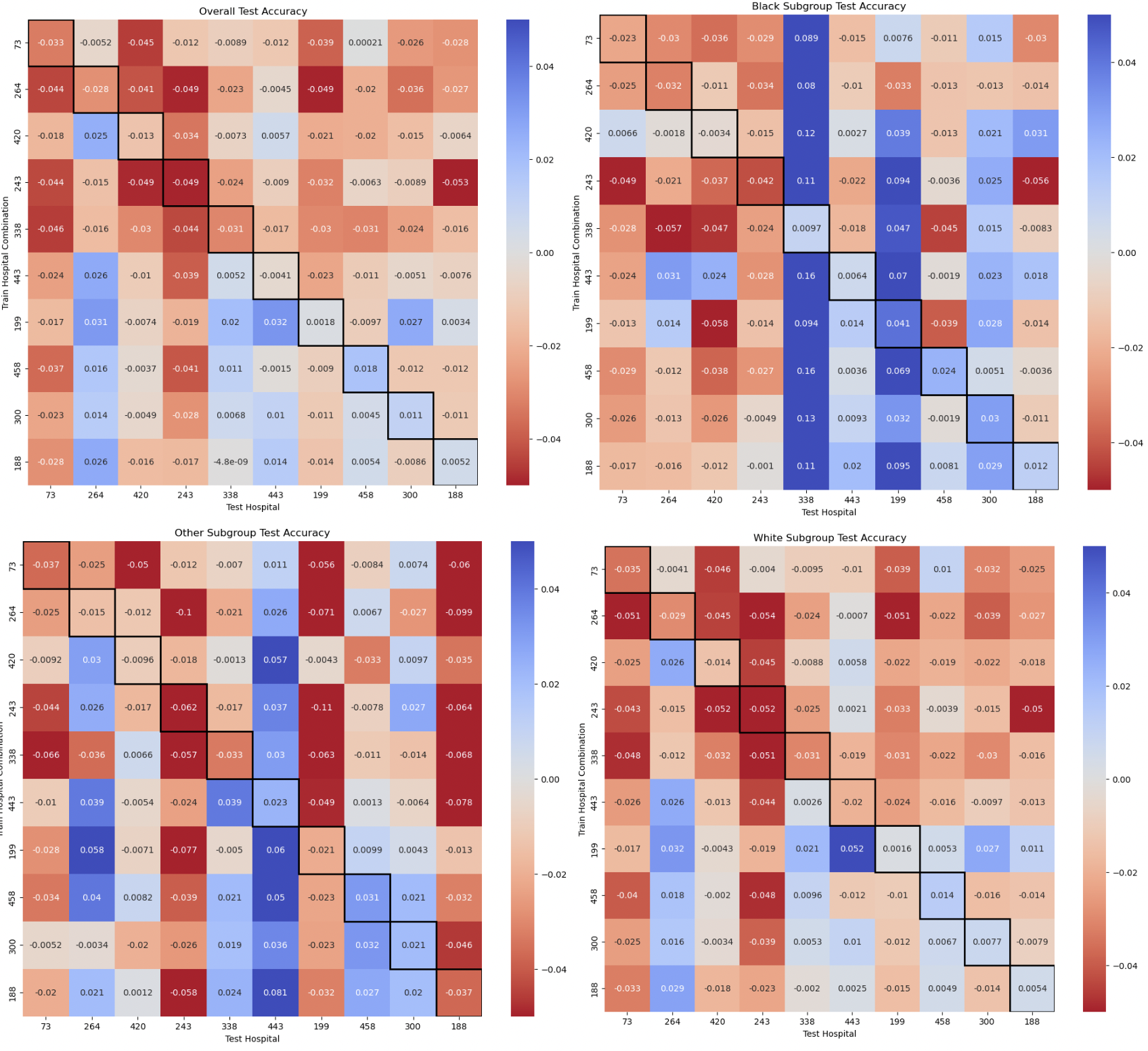}
    \caption{Change in Overall and Subgroup Accuracy after \textsc{Whole-Source} data addition (LSTM).}
    \label{fig:lstm-heatmaps}
\end{figure}

The full set of results for \textsc{Whole-Source} data addition using the Long Short-Term Memory (LSTM) model are presented in \ref{fig:lstm-heatmaps}. In general, we find that the Overall accuracy is harmed by any out-of-distribution data addition, more so than is seen with the LR or LGBM models. This is not a necessarily unexpected result, as LSTMs rely more heavily on learned sequential dependencies and higher-order feature representations. Regardless, we still observe that the effects on Overall accuracy are not equally reflected across all subgroups. For example, adding data to Test Hospitals 338 and 199 are greatly helpful for improving Black subgroup accuracy despite minimal or negative changes to Overall accuracy. Similarly, we observe improvements to the Other subgroup of Hospital 443 under data addition, but minimal to negative differences among all other groups.

\begin{figure}[h]
    \centering
    \includegraphics[width=0.7\linewidth]{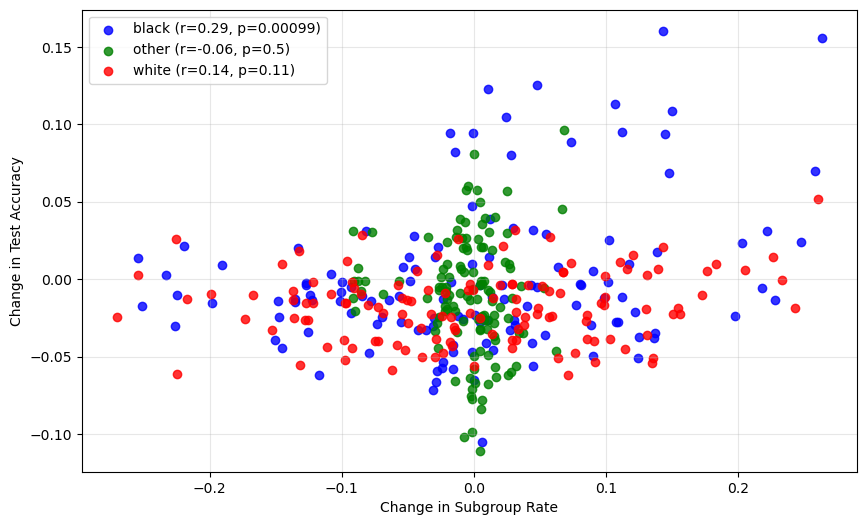}
    \caption{Change in Subgroup Rate vs. Change in Subgroup Accuracy after \textsc{Whole-Source} data addition, trained using an LSTM model.}
    \label{fig:lstm-subgroup-rate}
\end{figure}

\begin{figure}[h]
    \centering
    \includegraphics[width=1\linewidth]{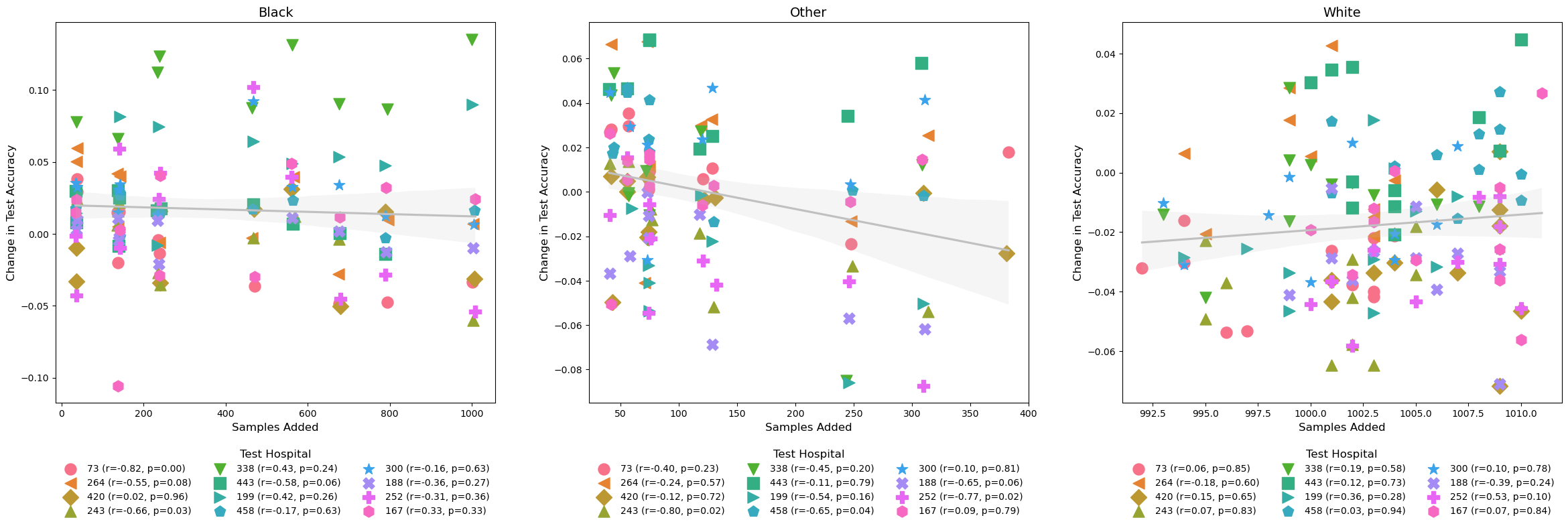}
    \caption{Number of subgroup samples added vs. Change in Subgroup Accuracy after \textsc{Subgroup-Level} data addition, trained using an LSTM model.}
    \label{fig:lstm-samples-added}
\end{figure}

\begin{figure}[h]
    \centering
    \includegraphics[width=1\linewidth]{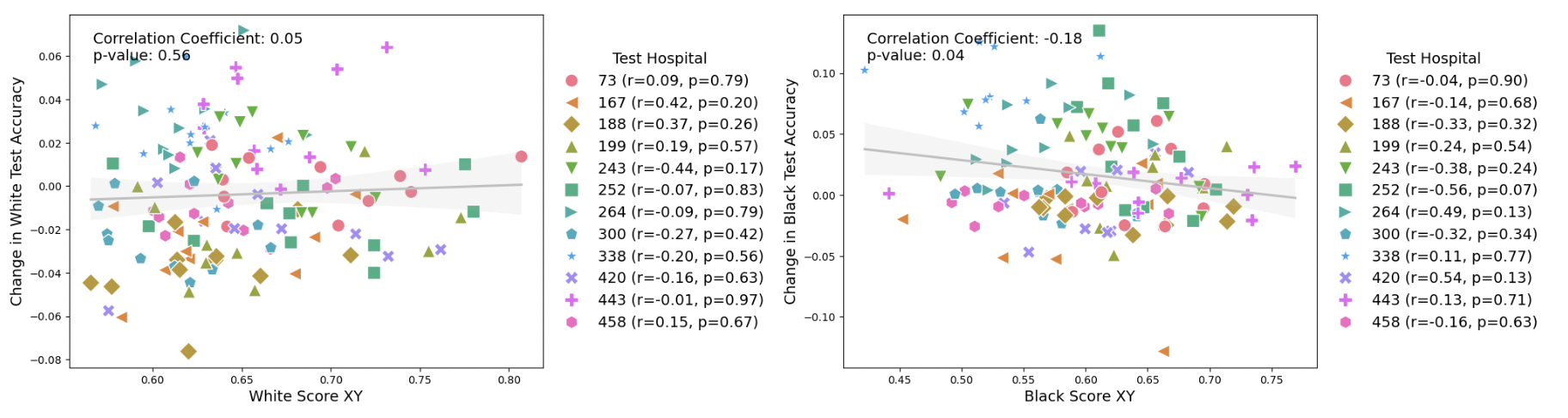}
    \caption{Subgroup Similarity Score between the test subgroup and added subgroup vs. Change in Subgroup Accuracy after \textsc{Subgroup-Level} data addition, trained using an LSTM model.}
    \label{fig:lstm-similarity-score}
\end{figure}

We applied the same three naive data addition methods as in Section \ref{sec:naive-data-addition}. The results are shown in Figures \ref{fig:lstm-subgroup-rate}, \ref{fig:lstm-samples-added}, and \ref{fig:lstm-similarity-score}. Interestingly, we observe that all three naive methods---increasing subgroup rate, sample size, and selecting the most similar sources---were informative for improving subgroup accuracy on the Black subgroup only. In each, we find a statistically significant relationship for the Black subpopulation. However, as these results were not seen consistently across all subgroups and model classes, we do not find that these methods overall are reliable strategies for improving subgroup performance on EHR datasets.

\end{document}